\documentclass[lettersize,journal]{IEEEtran}
\usepackage{amsmath,amsfonts}
\usepackage{algorithmic}
\usepackage{algorithm}
\usepackage{array}
\usepackage[caption=false,font=normalsize,labelfont=sf,textfont=sf]{subfig}
\usepackage{textcomp}
\usepackage{stfloats}
\usepackage{url}
\usepackage{verbatim}
\usepackage{graphicx}
\usepackage{wrapfig}
\usepackage{cite}
\usepackage[utf8]{inputenc} 
\usepackage[T1]{fontenc}    
\usepackage{url}            
\usepackage{booktabs}       
\usepackage{amsfonts}       
\usepackage{nicefrac}       
\usepackage{microtype}      
\usepackage{amsmath}
\usepackage{multirow}
\usepackage{multicol}
\usepackage[table,xcdraw]{xcolor}
\definecolor{lightgreen}{rgb}{0.8, 1, 0.8}
\usepackage{tikz}
\usepackage{float}
\usepackage{caption}

\begin{document}

\title{{\scshape ViKSeR}: Visual Knowledge-Driven Self-Reinforcing Reasoning Framework}

\author{
    Chao Wang,
    Chunbai Zhang,
    Yongxiao Tian,
    Yang Zhou,
    and Yan Peng
    \thanks{Chao Wang, Chunbai Zhang, Yongxiao Tian, and Yan Peng are with the School of Future Technology, Shanghai University, Shanghai, 200444, China.}%
    \thanks{Yang Zhou is with the School of Mechatronic Engineering and Automation, Shanghai University, Shanghai 200444, China.}%
    \thanks{Chao Wang is the corresponding author (e-mail: cwang@shu.edu.cn).}%
}

\markboth{}
{Chao Wang, Chunbai Zhang \MakeLowercase{\textit{et al.}}: ViKSeR: Visual Knowledge-Driven Self-Reinforcing Reasoning Framework}


\maketitle

\begin{abstract}
Visual reasoning refers to the task of solving questions about visual information.
Current visual reasoning methods typically employ pre-trained vision-language model~(VLM) strategies or deep neural network approaches.
However, existing efforts are constrained by limited reasoning interpretability, while hindering by the phenomenon of underspecification in the question text.
Additionally, the absence of fine-grained visual knowledge limits the precise understanding of subject behavior in visual reasoning tasks.
To address these issues, we propose {\scshape ViKSeR}~(\textbf{Vi}sual \textbf{K}nowledge-Driven \textbf{Se}lf-\textbf{R}einforcing Reasoning Framework).
Specifically, {\scshape ViKSeR}, trained using knowledge distilled from large language models, extracts fine-grained visual knowledge with the assistance of visual relationship detection techniques.
Subsequently, {\scshape ViKSeR} utilizes fine-grained visual knowledge to paraphrase the question with underspecification.
Additionally, we design a novel prompting method called Chain-of-Evidence~(CoE), which leverages the power of ``evidence for reasoning'' to endow {\scshape ViKSeR} with interpretable reasoning capabilities.
Meanwhile, the integration of self-reflection technology empowers {\scshape ViKSeR} with the ability to learn and improve from its mistakes.
Experiments conducted on widely used datasets demonstrate that {\scshape ViKSeR} achieves new state-of-the-art~(SOTA) results in relevant tasks.
Moreover, {\scshape ViKSeR} achieves performance on par with leading proprietary models, such as the latest ChatGPT-5.
\end{abstract}

\begin{IEEEkeywords}
 Chain-of-Evidence, Self-Reflection, Visual Knowledge, Visual Language Model, Visual Reasoning.
\end{IEEEkeywords}

\section{Introduction}\label{sec:intro}
\begin{figure}[!htbp]
\vskip 0.05in
\begin{center}
\centerline{\includegraphics[width=\columnwidth]{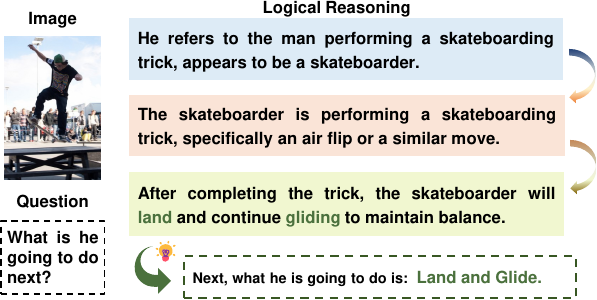}}
\caption{\textbf{A typical example of visual reasoning.} This task requires an agent to accurately align the image with the question and infer the intent through multi-step logical reasoning.}
\label{fig:introduce}
\end{center}
\vskip -0.25in
\end{figure}

Endowing machines with robust logical reasoning capabilities has been a long-standing goal of vision-language models~(VLMs)~\cite{paliGemma2024,llava12023,VRinVLM2021}.
A critical step toward realizing the goal lies in enhancing the model's visual reasoning capabilities~\cite{smola2024,cola2023}.
Visual reasoning involves solving questions about visual information~\cite{VPCVR2023,visualreasoning2021}, a task that necessitates precise alignment between visual and textual features, along with advanced logical reasoning skills~\cite{rapper2024,repare2024,VR2023}.
Figure~\ref{fig:introduce} illustrates a typical example of visual reasoning, where an agent must accurately align the image with the question and infer the intent through multiple steps of logical reasoning.
The notable features of visual reasoning not only drive advancements in cross-modal learning~\cite{molmo72b2024,vlmvqamedical2023} but also contribute to enhancing machines' logical reasoning abilities~\cite{repare2024,lingoqa2024}, underscoring its substantial research significance.

Existing visual reasoning methods typically employ pre-trained VLM strategies or deep neural network frameworks~\cite{paliGemma2024,molmo72b2024,VQAandVR2022}.
However, \textit{the reasoning capabilities of current efforts still exhibit limited interpretability} (i.e., \textbf{Issue 1}).
Additionally, \textit{underspecification} is a common phenomenon in visual reasoning tasks~\cite{repare2024, specification2023}, where a ambiguous description of the subject in the question text can hinder the alignment between textual and visual features, \textit{leading to multiple incorrect visual interpretations and undermining reasoning reliability} (i.e., \textbf{Issue 2}).
As shown in Figure~\ref{fig:introduce}, the subject ``he'' in the question is ambiguously described, which may be confused with other individuals in the image, presenting a risk of incorrect reasoning.
On the other hand, visual knowledge extraction (VKE) techniques, which provide enriched visual information, have the potential to significantly assist in addressing visual reasoning tasks~\cite{openvik2024,ivlr2024}. 
However, \textit{existing VKE methods are insufficient in extracting latent and fine-grained visual knowledge} (i.e., \textbf{Issue 3}).
The issue limits the accurate understanding of subject behavior in complex questions, thereby hindering the interpretability and precision required for visual reasoning tasks.

To address these issues, we design {\scshape ViKSeR}~(\textbf{Vi}sual \textbf{K}nowledge-Driven \textbf{Se}lf-\textbf{R}einforcing Reasoning Framework). 
The core components of {\scshape ViKSeR} are the fine-grained visual knowledge extraction~(\texttt{F-VKE}) module and the self-reinforcing reasoning~(\texttt{S-RR}) module.
For \textbf{Issue 1}, we deploy the \texttt{S-RR} module, which offers enhanced interpretability in reasoning and integrates self-reinforcement capabilities.
Specifically, the module integrates a novel prompting technique, Chain-of-Evidence~(CoE), which we propose to enhance interpretability by facilitating step-by-step reasoning grounded in factual evidence.
Meanwhile, a self-reflection mechanism, which utilizes insights from past failures to refine future reasoning, is introduced to facilitate adaptive reinforcement.
Additionally, intending to address \textbf{Issue 2},  a specification paraphrase method is explored for the \texttt{S-RR} module.
Concretely, we collect object-level information from fine-grained visual knowledge to refine ambiguous descriptions of the subject in question, thereby improving the question's clarity and completeness.

To address \textbf{Issue 3}, we deploy the \texttt{F-VKE} module to provide fine-grained visual knowledge to {\scshape ViKSeR}.
To be specific, the \texttt{F-VKE} module first detects the detailed visual relationships among key entities in an input image.
Subsequently, we train the \texttt{F-VKE} module using knowledge distilled from large language models (LLMs) to uncover causal relationships between entities' behaviors and their inferred outcomes. For instance, as shown in Figure~\ref{fig:kdAncase}, LLMs can infer that a man depicted squatting on the right side of the train door may disembark at the next station.
Following this, the trained \texttt{F-VKE} module leverages visual and causal relationships to generate fine-grained visual knowledge.
A specific case that validates the fine-grained nature and richness of the visual knowledge generated by the \texttt{F-VKE} module is provided in Section~\ref{subsec:case study}.
We conduct extensive experiments on diverse widely-recognized datasets to validate {\scshape ViKSeR}'s visual knowledge extraction and reasoning capabilities. The results demonstrate that {\scshape ViKSeR} outperforms the latest research across all datasets, achieving new SOTA results.
We summarize our \textbf{contributions} as follows:
\begin{itemize}
\item We propose {\scshape ViKSeR}, a novel framework for visual reasoning tasks, which extracts fine-grained and enriched visual knowledge while performing highly interpretable and self-reinforcing reasoning.
\item A specification paraphrase method is designed to help {\scshape ViKSeR} mitigate underspecification, while a novel CoE prompting technique and a self-reflection mechanism are introduced to assist {\scshape ViKSeR} in performing highly interpretable and self-reinforcing reasoning. 
\item Extensive experimental results demonstrate {\scshape ViKSeR}’s improvements over advanced baselines across diverse public datasets, achieving new SOTA results.
\end{itemize}

\section{Related Work} \label{sec:related}
\subsection{Visual Knowledge Extraction}
Existing VKE methods either employ a holistic image captioning strategy or rely on fixed knowledge formats to extract visual information~\cite{PV2TEA2023,llava12023}.
Specifically, deep learning has played a crucial role in advancing image captioning approaches~\cite{DLimgcap2018,DLimgcap2017}.
With the development of VLMs, an increasing number of studies leverage pre-trained multimodal large language models~(MLLMs) to understand both visual and textual information in a unified framework~\cite{promptcap2023,ISCimgcap2022}. 
On the other hand, the image captioning methods provide a broad understanding of the image content, yet they encounter challenges in conveying fine-grained details.
In contrast, some studies utilize fixed knowledge graphs to map visual features to predefined semantic categories~\cite{ivlr2024,VCHG2023}, providing a structured form of visual knowledge~\cite{covlm2024,BKG2GSG2020}. While these methods ensure consistency and interpretability, they fall short in providing enriched visual knowledge.

\subsection{Language Model Reasoning}
Recently, large VLMs have demonstrated remarkable success in reasoning and inference, particularly in facilitating few-shot~\cite{molmo72b2024, cotrainfewshot2022} and zero-shot~\cite{llm0shottrain2024,llm0shot2022} learning. 
Notably, through specialized training on diverse cross-modal benchmarks, VLMs have demonstrated advanced visual comprehension and reasoning abilities~\cite{smola2024,paliGemma2024,palix2022}, significantly enhancing performance in visual reasoning tasks.
Additionally, the potential of prompt-based reasoning~\cite{repare2024,rapper2024,openvik2024} has been extensively explored to tackle diverse cross-modal visual reasoning tasks, including visual question answering~(VQA)~\cite{repare2024,smola2024}, visual entailment~(VE)~\cite{esnlive2021}, and visual commonsense reasoning~(VCR)~\cite{rapper2024,nlxgpt2022}. 
By leveraging the substantial information embedded within VLMs, novel insights are generated to advance reasoning and interpretability research.
In this paper, we aspire to employ fine-grained visual knowledge as factual grounding while implementing our interpretable, self-reinforcing reasoning paradigm to enhance performance on visual reasoning tasks.

\section{Method}\label{sec:methods}
\begin{figure*}[t]
\begin{center}
\centerline{\includegraphics[width=\textwidth]{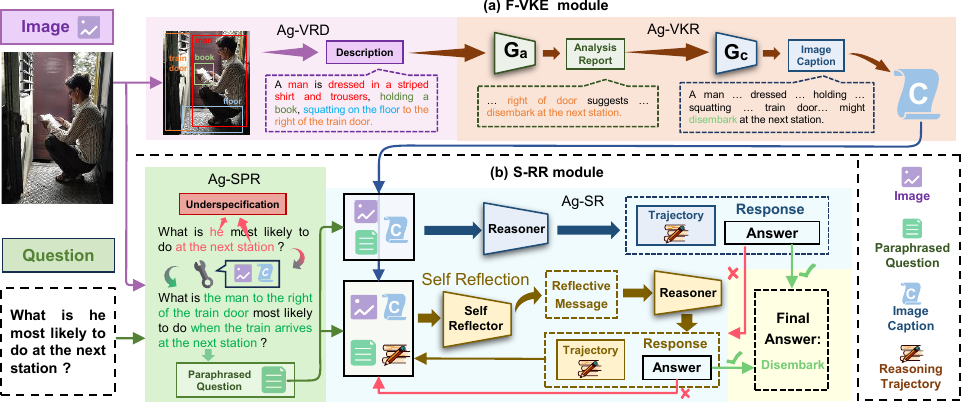}}
\caption{\textbf{The framework of {\scshape ViKSeR}}. Starting with the input image and question, {\scshape ViKSeR} first extracts the relationships between key entities in the image and analyzes these relationships to form a detailed image caption. Then, {\scshape ViKSeR} uses the image caption to paraphrase the question, refining the ambiguous
descriptions in the question. Finally, {\scshape ViKSeR} infers a predicted answer based on the paraphrased question, the image, and its caption. The predicted answer is taken as the final answer if it sufficiently addresses the question. Otherwise, {\scshape ViKSeR} performs self-reflection based on the reasoning trajectory, the paraphrased question, the image, and its caption to seek a new response.}
\label{fig:framework}
\end{center}
\vskip -0.2in
\end{figure*}

In this section, we provide detailed descriptions of {\scshape ViKSeR}'s \texttt{F-VKE} module (Section~\ref{subsec:FVKE}) and \texttt{S-RR} module (Section~\ref{subsec:SRR}).
Figure~\ref{fig:framework} illustrates a detailed framework of {\scshape ViKSeR}. Next, we will provide a detailed introduction to the modules of {\scshape ViKSeR}.

Specifically, the \texttt{F-VKE} module consists of two agents: the Visual Relationship Detector~(\textit{Ag-VRD}) and the Visual Knowledge enRicher~(\textit{Ag-VKR}), where \textit{Ag-VKR} consists of a causal relationship analyzer $ G_a $ and an image caption generator $ G_c $.
Starting with the input image $ I $ and question $ Q $, \textit{Ag-VRD} first detects the visual relationships between key entities in $ I $ and integrates these relationships into a preliminary visual description $ D $.
Following this, $ G_a $ in \textit{Ag-VKR} provides an analysis report $ A $ of the causal relationships between key entities' behaviors in $ D $ and their inferred outcomes.
For instance, a description of a man squatting at the train door implicitly suggests that he may disembark at the next station.
Finally, leveraging $ A $, $ G_c $ in \textit{Ag-VKR} enriches $ D $ into a more detailed and nuanced image caption $ C $, offering fine-grained visual knowledge.
On the other hand, the \texttt{S-RR} module is composed of the specification paraphraser~(\textit{Ag-SPR}) and the self-refining reasoner~(\textit{Ag-SR}). 
Specifically, \textit{Ag-SPR} first collects visual information from $ C $ to refine ambiguous descriptions of the subject in $ Q $, thereby forming the paraphrased question $ Q_r $.
Subsequently, \textit{Ag-SR} employs CoE prompting to infer the predicted answer $ \tilde{a} $ based on $ Q_r $, $ I $, and $ C $. When $ \tilde{a} $ exhibits positivity, it directly serves as the final answer $ a $. Conversely, if $ \tilde{a} $ shows negativity, \textit{Ag-SR} will use $ \tilde{a} $ and the reasoning trajectory $ t $ to perform self-reflection in search of a more valuable response. 
For the sake of readability, the key notations and their meanings used in this paper are systematically summarized in Table~\ref{tab:notations}. 

\begin{table}[h]
\caption{Comprehensive overview of the key notations and their meanings used in this paper.}
\vskip -0.1in
\label{tab:notations}
\begin{center}
\begin{small}
\begin{tabular}{cp{7cm}}
\toprule
\textbf{Notation} & \textbf{Meaning}\\
\midrule 
$I$ & The input image\\
$Q$ & The original question \\
$E_K$ & The key entity\\
$r$ & The visual relationship between $E_K$\\
$R_K$ & The key visual relationship\\
$D$ & The preliminary visual description\\
$C$ & The detailed image caption\\
$A$ & The analysis report about causal relationships between key entities’ behaviors and their inferred outcomes\\
$A_p$ & The pseudo-ground-truth causal relationship analysis report used in the knowledge distillation process\\
$C_p$ & The pseudo-ground-truth image caption used in the knowledge distillation process\\
$G_a$ & The causal relationship analyzer\\
$G_c$ & The image caption generator\\
$Q_r$ & The paraphrased question\\
$\tilde{a}$ & The predicted answer to the question\\
$a$ & The final answer to the question\\
$t$ & The reasoning trajectory\\
$S_{ref}$ & The discrete binary reward score assigned to $\tilde{a}$\\
$V_{ref}$ & The verbal reflective message generated by the self-reflection mechanism\\
\bottomrule
\end{tabular}
\end{small}
\end{center}
\vskip -0.3in
\end{table}

\subsection{F-VKE Module}\label{subsec:FVKE}
Guided by the principles of multi-agent collaboration, the functionality of the \texttt{F-VKE} module in extracting fine-grained visual knowledge is distributed across two agents: the Visual Relationship Detector \textit{Ag-VRD} and the Visual Knowledge enRicher \textit{Ag-VKR}.
The architectural design of \textit{Ag-VRD} and the \textit{Ag-VKR} is elaborated as follows.

\paragraph{\textbf{Visual Relationship Detector}}\label{subsubsec:VRD}
Accurate detection of entities in images and their relationships form the foundation for effective VKE techniques. 
With the aim of endowing the \texttt{F-VKE} module with efficient visual relationship detection capabilities, we design the \textit{Ag-VRD} based on existing VRD methods.
Concretely, given an input image $I$, the \textit{Ag-VRD} is prompted to extract all entities within the image.
Each entity is then assigned a validity score $S_e$, reflecting its relevance in understanding the image's content.
Notably, when the entity's score exceeds a predefined threshold $\theta_e$, it is classified as a key entity $E_K$.
Following this, with the aid of CoVLM~\cite{covlm2024}, we extract all relevant visual relationships of $E_K$. 
CoVLM is an exceptional VRD method that facilitates the collaboration between visual detection networks and LLMs, utilizing specially designed communication tokens. The advantage of CoVLM lies in its ability to precisely identify the target entity $E_T$ corresponding to a given relationship based on the input entity $E_I$, by leveraging the specially crafted communication tokens.
However, due to the fixed structure of its communication tokens, CoVLM is limited in offering fine-grained visual knowledge. Therefore, we utilize CoVLM primarily for refining the detected visual relationships.

To be specific, we first employ the detection network of CoVLM to identify the regions of interest corresponding to each $E_K$ in $ I $. 
Based on these regions and their associated entities, the \textit{Ag-VRD} is then prompted to detect all potential relationships associated with $E_K$.
Subsequently, we employ CoVLM to detect the relevant entities corresponding to these potential relationships. 
These potential relationships are then merged with their associated entities to form the visual relationship $ r $.
Following this, the \textit{Ag-VRD} is employed to assess a relationship validity score $ S_r $ for each $ r $ based on its effectiveness in contributing to the understanding of $ I $.
To determine which visual relationships associated with the key entities are truly effective in capturing the semantics of the input image, we propose a joint entity-relation validity evaluation algorithm to compute the entity-relation joint validity score $ S_r^e $:
\begin{equation}
S_r^e = S_e \cdot \omega_r \cdot S_r, \omega_r = 1 + \gamma (\alpha - N_r^e),
\label{eq:jointscore}
\end{equation}
where $S_e$ denotes the entity validity score and $S_r$ denotes the relationship validity score. $ \omega_r $ represents the weight associated with the current visual relationship $ r $. $ N_r^e $ denotes the number of $ r $ held by the key entity $ E_K $. $ \gamma $ and $ \alpha $ are hyperparameters, where $ \gamma $ controls the influence of the number of visual relationships on the weight, constrained within the range $ (0.05, 0.2) $. And $ \alpha $ limits the number of visual relationships.
For example, as $ N_r^e $ increases, $ \omega_r $ decreases to amplify the constraint of $ S_r^e $ on the visual relationship. Conversely, $ \omega_r $ increases when the number of relationships is smaller.
Additionally, in case $ S_r^e $ exceeds a threshold $ \theta_r^e $, the corresponding visual relationship under the current key entity is labeled as a key visual relationship $R_K$.
After identifying $E_K$ and $R_K$, we generate the preliminary image description $ D $ by progressively aligning each $E_K$ with its corresponding key visual relationship $R_K$.

\paragraph{\textbf{Visual Knowledge Enricher}}\label{subsubsec:VKR}
Existing VRD methods primarily focus on detecting surface-level, concept-based relational information within an image~\cite{covlm2024, spatialRelation2024}. 
This nature makes it challenging for current VRD methods to conduct in-depth, object-level analysis of the image content. 
However, such analyses are essential for a comprehensive understanding of the image.
For this reason, we further deploy \textit{Ag-VKR}, which consists of a causal relationship analyzer $ G_a $ and an image caption generator $ G_c $, to enhance the depth of knowledge in image analysis.
Specifically, $ G_a $ first interprets the preliminary image description $ D $ concerning the input image $ I $. 
Next, $ G_a $ derives an analysis report $ A $ that uncovers the causal relationships between the behaviors of key entities in $ D $ and their inferred outcomes.
Subsequently, $ G_c $ enriches $ D $ by incorporating the visual knowledge from $ I $ and $ A $, resulting in a detailed image caption $ C $ for $ I $.

\begin{figure}[h]
    \begin{center}
    \centerline{\includegraphics[width=\columnwidth]{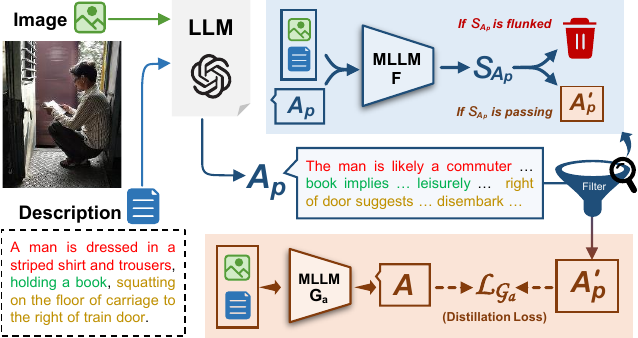}}
    \caption{Process of utilizing knowledge distillation from LLM to train $G_a$ to generate $A$.}
    \label{fig:kdAncase}
\end{center}
\vskip -0.25in
\end{figure}

Causal relationship analysis reporting and detailed image captioning typically require training with ground-truth data, which is often absent in existing datasets.
To address this, we propose leveraging the extensive knowledge reservoir inherent in LLMs (e.g., ChatGPT-4o~\cite{gpt42023}) to generate causal relationship analysis reports and detailed image captions as pseudo-ground-truth data to train $ G_a $ and $ G_c $.
Specifically, we extract pseudo-ground-truth causal relationship analysis report $ A_p $ from LLM with a task-specific set of few-shot demonstrations as follows:
\begin{equation}
    KA_p = \left \{ A_p \ \middle| \ A_p \sim P_{LLM}(I, D) \right \},
    \label{eq:KDKAp}
\end{equation}
where $ I $ denotes the input image, $ D $ represents the ground-truth preliminary image description of $ I $, and $ P_{LLM} $ denotes the LLM operating in an autoregressive manner. $ A_p $ represents the pseudo-ground-truth causal relationship analysis report sampled from $ P_{LLM} $, and $ KA_p $ represents the set of all $ A_p $.
Notably, $ KA_p $ may be noisy and erroneous, which could adversely affect the subsequent detailed image captioning process.
To address this, we apply a post-processing mechanism to filter $ KA_p $ into $ KA'_p $.
Specifically, for each $ A_p $ in $ KA_p $, we use a pre-trained MLLM (e.g., ChatGPT-4o; denoted as $F$) to assess its validity score $ S_{A_p} $ based on whether $ A_p $ correctly uncovers the causal relationship. In case $ S_{A_p} $ exceeds a predetermined threshold $ \tau $, the corresponding $ A_p $ is retained. The process of collecting $ KA’_p $  is formalized as follows:
\begin{equation}
    KA'_p = \left\{ A_p \ \middle| \ S_{A_p} > \tau \right\}, S_{A_p} = F\left(A_p, (I, D)\right),
    \label{eq:KAprefine}
\end{equation}
where $ A_p $ denotes the pseudo-ground-truth causal relationship analysis report. $ I $ denotes the input image, $ D $ represents the ground-truth preliminary image description of $ I$. $\tau $ denotes the predetermined threshold. $F$ denotes the pre-trained MLLM.
With $ KA’p $ serving as pseudo-ground-truth data, we are able to train $ G_a $ with the distillation loss $ L_{G_a} $, as formalized below:
\begin{equation}
    L_{G_a} = - \sum_{t=1}^{T} \log \left[ P_{G_a} \left( A_{p,t}' \ \middle| \ A_{p,t-1}', (I,D) \right) \right],
    \label{eq:GaKD}
\end{equation}
where $ A’_p \in KA’_p $, and $ T = |A’_p| $. Figure~\ref{fig:kdAncase} shows how we utilize knowledge distillation from the LLM to train $ G_a $.
Specifically, for the Image and Description, a post-processing mechanism employs $ F $ to determine whether $ S_{A_p} $ is passing, thereby filtering out the noise from the LLM-generated $ A_p $ to obtain $ A'_p $.
Following this, $ A'_p $ is used to train $ G_a $ via $ L_{G_a} $, in order to generate $ A $.

Similarly, we employ the same knowledge distillation approach to train $ G_c $. Concretely, we utilize $ KA’_p $ as input to help $ G_c $ enrich the preliminary image description $ D $, thereby generating pseudo-ground-truth image captions $ KC’_p $. Subsequently, we train the image caption generator $ G_c $ with the distillation loss $ L_{G_C} $ described below:
\begin{equation}
    L_{G_c} = - \sum_{t=1}^{T} \log \left[ P_{G_c} \left( C_{p,t}' \ \middle| \ C_{p,t-1}', (I, D, A'_p) \right) \right],
    \label{eq:GcKD}
\end{equation}
where $ C’_p \in KC’_p $, $ A’_p \in KA’_p $, and $ T = |C’_p| $. More details are discussed in Appendix~\ref{apd:trainGc} and \ref{apd:KDfromLLM}.
In summary, the trained causal relationship analyzer $ G_a $ and image caption generator $ G_c $ operate collaboratively to fulfill the functionality of \textit{Ag-VKR}. Specifically, given an input image $ I $ and the preliminary image description $ D $, $ G_a $ and $ G_c $ generate a detailed image caption $ C $ for $ I $.

\subsection{S-RR Module}\label{subsec:SRR}
We leverage the detailed visual knowledge in the image caption $ C $, which is generated by the \texttt{F-VKE} module, to enhance the performance of {\scshape ViKSeR} in visual reasoning tasks, such as VQA and VE.
To achieve this, we propose an \texttt{S-RR} module, consisting of a specification paraphraser \textit{Ag-SPR} and a self-refining reasoner \textit{Ag-SR}, to endow {\scshape ViKSeR} with advanced reasoning capabilities.
Specifically, starting with an image and a question, \textit{Ag-SPR} extracts and analyzes the fine-grained visual knowledge in $ C $, to refine ambiguous descriptions in the question and paraphrase it. 
Subsequently, \textit{Ag-SR} resolves the paraphrased question by leveraging the input image and the detailed image caption through a self-reinforcing reasoning paradigm.
It is important to emphasize that we aim to develop a reasoning approach characterized by high generalizability and a gradient-free, in light of the diversity of visual reasoning tasks.. For this reason, we propose employing a flexible and plug-and-play reasoning prompt paradigm to prompt pre-trained visual language models~(PVLMs) to perform the functions of \textit{Ag-SPR} and \textit{Ag-SR}. Detailed prompts are discussed in Appendix~\ref{apd:prompts}. The architectural design of \textit{Ag-SPR} and the \textit{Ag-SR} is elaborated as follows.

\paragraph{\textbf{Specification Paraphraser}}\label{subsubsec:specification paraphraser}
Starting with the input image $ I $, the original question $ Q $, and the detailed image caption $ C $ generated by the \texttt{F-VKE} module, \textit{Ag-SPR} leverages the detailed visual information in $ C $ to paraphrase $ Q $, thereby refining any ambiguous descriptions within $ Q $. We employ a PVLM as the core of \textit{Ag-SPR} to facilitate this process.
Specifically, we prompt the PVLM in \textit{Ag-SPR} to first identify the main subject in $ Q $, and then query the detailed image caption $ C $ for relevant textual information. Meanwhile, the PVLM is prompted to extract the interaction between the main subject and the image scene from $ C $, providing a complementary description of the main subject's context. Subsequently, the above textual information is paraphrased in natural language and integrated into $ Q $ to form the paraphrased question $ Q_r $. The underlying logic of this process is that the entities mentioned in $ Q $ intuitively provide key information relevant to the expected answer intuitively.
For this reason, providing more detailed descriptions of the subject aids the \texttt{S-RR} module in resolving the visual reasoning task.

\paragraph{\textbf{Self-Refining Reasoner}}\label{subsubsec:selfrefining reasoner}
As a key component of the \texttt{S-RR} module, the cognitive and reasoning capabilities of \textit{Ag-SR} significantly influence the overall problem-solving performance of the system. To enhance \textit{Ag-SR}'s capabilities, we introduce a prompting technique called CoE, which guides \textit{Ag-SR} to reason incrementally and provide highly interpretable steps toward the expected answer of $ Q_r $. Furthermore, we present a self-reflection mechanism that allows \textit{Ag-SR} to perform self-correction, thereby improving its robustness in reasoning. Detailed descriptions are as follows.

\textbf{Highly Interpretable Reasoning}: We prompt a PVLM to serve as the core of \textit{Ag-SR}.
Given that conventional reasoning prompts, such as ``think step-by-ste''~\cite{COT2022}, often induce hallucinations due to insufficiently grounded rationale, we propose CoE, a tailored prompting technique, to enhance the PVLM’s reasoning ability.
Concretely, the PVLM first analyzes fine-grained visual knowledge provided by the \texttt{F-VKE} module, extracting factual information to serve as evidence for reasoning.
Subsequently, CoE guides the PVLM through a structured, step-by-step reasoning process based on the extracted evidence, ultimately generating a predicted answer $ \tilde{a} $. 
This reasoning framework enables CoE to enhance interpretability while improving reasoning accuracy.
Further details on CoE prompting are provided in Appendix~\ref{apd:coeprompt}.
Additionally, we validate the effectiveness of the CoE prompting method in Section~\ref{subsec:ablation studies}.

\textbf{Self-Reflection-Based Reinforcement Mechanism}: Existing research has shown that even highly intelligent agents are susceptible to generating low-quality answers, which significantly diminishes the accuracy of their reasoning and decision-making processes.
To address this, we propose a self-reflection mechanism that builds upon the method proposed by Shinn et al~\cite{reflexion2023}. The mechanism enables \textit{Ag-SR} to manage instances of low-quality answers by leveraging past experiences, thereby reducing the likelihood of similar errors occurring in the future.
Specifically, after deriving $ \tilde{a} $, \textit{Ag-SR} utilize an exact matching mechanism to map complex natural language information in $ \tilde{a} $ to a two-dimensional discrete binary reward score $ S_{ref} $. 
If $ S_{ref} $ is positive, the predicted answer $ \tilde{a} $ is directly used as the final answer $ a $. 
In contrast, when $ S_{ref} $ is represented as negative, \textit{Ag-SR} is prompted to analyze $ Q_r $ and a reasoning trajectory $ t $ from the previous trial. Subsequently, \textit{Ag-SR} reflects on the cause of failure and generates a detailed verbal reflection, denoted as $ V_{ref} $. 
Following this, $ V_{ref} $ serves as experiential knowledge to guide \textit{Ag-SR} in generating a more accurate answer in subsequent trials.
It is important to note that, compared to the reward score $ S_{ref} $, $ V_{ref} $ encompasses more comprehensive experiential information. For instance, while $ S_{ref} $ merely indicates success or failure, $ V_{ref} $ provides insight into which aspects of the previous reasoning trajectory were incorrectly applied, resulting in previous failure. Furthermore, $ V_{ref} $ can provide guidance on how to avoid similar mistakes in future trials.
Further details are discussed in Appendix~\ref{apd:selfref}.

\begin{table*}[t]
\caption{Comparative performance of different methods on the VQAv2, A-OKVQA, and VizWiz datasets. \raisebox{0.5ex}{\colorbox{lightgreen}{\makebox[1em]{}}} represents the optimal scores, while \raisebox{0.5ex}{\colorbox{yellow!30}{\makebox[1em]{}}} represents the suboptimal scores. We replicate the scores of the baselines marked with \textsuperscript{$\dagger$}, while the scores for the remaining baselines are sourced from the official repositories of VQAv2, A-OKVQA, and VizWiz datasets, as well as the study conducted by Prasad et al.~\cite{repare2024}.}
\label{tab:vqa task}
\begin{center}
\begin{small}
\begin{tabular}{p{3.1cm}|>{\centering\arraybackslash}p{1.5cm}>{\centering\arraybackslash}p{1.5cm}>{\centering\arraybackslash}p{1.5cm}>{\centering\arraybackslash}p{1.5cm}|>{\centering\arraybackslash}p{1.5cm}>{\centering\arraybackslash}p{1.5cm}|>{\centering\arraybackslash}p{1.5cm}}
\toprule
\multirow{2}{*}{\textbf{Methods}} & \multicolumn{4}{c|}{\textbf{VQAv2}} & \multicolumn{2}{c|}{\textbf{A-OKVQA}} & \textbf{VizWiz} \\
\cmidrule(lr){2-5} \cmidrule(lr){6-7} \cmidrule(lr){8-8}
& Ov.~(\%) & Y/N~(\%) & Num.~(\%) & Ot.~(\%) & MC.~(\%) & DA.~(\%) & Ov.~(\%) \\
\midrule
\multicolumn{8}{c}{\textit{General Methods}} \\
LLaVa-1.5-7B\textsuperscript{$\dagger$} & 74.04 & 86.84 & 52.94 & 57.89 & 62.56 & 77.38  & 57.07\\
REPARE+LLaVa-1.5 & 77.34 & 92.64 & 59.92 & 70.12 & 66.19 & 78.21 & 59.46 \\
REPARE+BLIP-2 & 74.05 & 94.56 & 54.40 & 63.36 & 55.67 & 82.80 & 70.03 \\
Molmo-72B & 86.67 & 96.91 & \cellcolor{lightgreen}\textbf{78.49} & 79.63 & 82.20 & 74.95 & - \\
ChatGPT-4o-mini\textsuperscript{$\dagger$} & 75.54 & 81.69 & 65.46 & 76.41 & 82.76 & 78.79 & 80.61 \\
ChatGPT-4o\textsuperscript{$\dagger$} & 86.98 & 93.44 & 71.43 & \cellcolor{yellow!30}85.94 & 90.79 & 89.79 & 79.87 \\
ChatGPT-5\textsuperscript{$\dagger$} & \cellcolor{yellow!30}88.68 & 93.86 & 77.86 & 85.33 & \cellcolor{yellow!30}91.86 & \cellcolor{lightgreen}\textbf{90.47} & \cellcolor{yellow!30}82.15 \\

\midrule
\multicolumn{8}{c}{\textit{VRD Methods}} \\
CoVLM\textsuperscript{$\dagger$} & 48.80 & 64.02 & 36.85 & 40.37 & 46.47 & 52.35 & 44.32\\
\midrule
\multicolumn{8}{c}{\textit{Task-Specific Methods}} \\
MAVL & - & - & - & - & 53.80 & 50.70 & - \\
DIETCOKE & - & - & - & - & 49.20 & 48.60 & - \\
LocVLM-L & 55.90 & 68.73 & 40.45 & 46.37  & - & - & - \\
HiMix & 80.20 & 89.72 & 61.37 & 73.48 & - & - & -\\
CIBi & 61.59 & 82.02 & 43.60 & 51.79 & - & - & - \\
SMoLA-PaLI-X & 85.00 & - & - & - & 84.1 & 70.55 & 72.00 \\
PaLI-X & 86.06 & 96.78 & 74.14 & 79.46 & 80.4 & 68.20 & 73.30 \\
PaLI-Gemma 2 & 86.95 & \cellcolor{yellow!30}97.19 & 77.77 & 80.13 & 83.7 & 71.30 & 78.10 \\
\midrule
{\scshape ViKSeR}~(ours) & \cellcolor{lightgreen}\textbf{88.74} & \cellcolor{lightgreen}\textbf{98.31} & \cellcolor{yellow!30}75.51 & \cellcolor{lightgreen}\textbf{86.71} & \cellcolor{lightgreen}\textbf{92.51} & \cellcolor{yellow!30}88.53 & \cellcolor{lightgreen}\textbf{82.52} \\
\bottomrule
\end{tabular}
\end{small}
\end{center}
\vskip -0.2in
\end{table*}

\section{Experiments}\label{sec:experiments}
In this section, we conduct extensive experiments on publicly available datasets to evaluate {\scshape ViKSeR}'s performance on visual reasoning tasks.

\subsection{Experimental Setup}\label{subsec:experimental setup}
\textbf{Datasets:~}To evaluate the reasoning abilities of {\scshape ViKSeR}, we utilize widely-used datasets for visual reasoning tasks, including VQAv2~\cite{vqav22017}, A-OKVQA~\cite{aokvqa2022}, VizWiz~\cite{vizwiz2018}, e-SNLI-VE~\cite{esnlive2021}, COLA~\cite{cola2023}, and CREPE~\cite{crepe2023}.

\textbf{Baselines:~}We compare {\scshape ViKSeR} with \textbf{22} competitive baselines, including SMoLA-PaLI-X~\cite{smola2024}, LLaVa-1.5-7B~\cite{llava12023}, Qwen-2.5~\cite{qwen2.5-VL}, ChatGPT-4o-mini~\cite{4omini2024}, ChatGPT-4o~\cite{gpt4o2024}, ChatGPT-5~\cite{gpt52025}, REPARE~\cite{repare2024}, CoVLM~\cite{covlm2024}, Molmo-72B~\cite{molmo72b2024}, PaLi-X~\cite{palix2022}, PaliGemma 2~\cite{paliGemma2024}, NLX-GPT~\cite{nlxgpt2022}, LOCVLM-L~\cite{LocVLM2024}, HIMIX~\cite{HiMix2025}, CIBI~\cite{CIBi2024}, MAVL~\cite{MAVL2024}, DIETCOKE~\cite{DIETCOKE2024}, Rapper~\cite{rapper2024}, e-UG~\cite{esnlive2021}, OFX-X~\cite{ofx2022}, MosaiCLIP~\cite{mosai2023}, CLIP+ MM-Pred~\cite{cola2023}, and  CLIP+ Linear~\cite{cola2023}.

\textbf{Evaluation Metrics:~}
Based on previous work~\cite{cola2023,crepe2023,repare2024,rapper2024}, we evaluate accuracy across distinct question types: yes/no~(Y/N), number~(Num.), other~(Ot.), direct answer~(DA.), and multiple-choice~(MC.), as well as the overall accuracy~(Ov.).
Notably, for the COLA, CREPE, and e-SNLI-VE datasets, we report Accuracy (Acc.) directly. Additionally, for CREPE, e further evaluate Recall@1~(R@1).

\textbf{Implementation Details}
We prompt GPT-4o mini~\cite{4omini2024} as the LLM foundation for \textit{Ag-VRD} of the \texttt{F-VKE} module in {\scshape ViKSeR}.
Additionally, we employ LLaVa-1.5-7B~\cite{llava12023} as the PVLM foundation for the \texttt{S-RR} module of {\scshape ViKSeR}. 
It is important to note that in Section~\ref{subsubsec:VKR}, the LLM used for knowledge distillation is ChatGPT-4o~\cite{gpt42023}, while $ G_a $ and $ G_c $ are obtained through training LLaVa-1.5-7B.
On the other hand, a summary of the experimental setup is provided, which includes the datasets, baseline methods, and evaluation metrics employed for performance assessment.

\subsection{Main Results}\label{subsec:mainresults}
We conduct extensive experiments to validate {\scshape ViKSeR}'s visual reasoning capabilities, with the main experimental results summarized as follows.

\textbf{{\scshape ViKSeR} performs outstandingly in reasoning and answering visual questions.~}To validate the superiority of {\scshape ViKSeR}'s visual reasoning capabilities, we compare its performance with ten competitive baselines on the VQAv2, A-OKVQA, and VizWiz datasets, as presented in Table~\ref{tab:vqa task}.
To be specific, {\scshape ViKSeR} achieves new SOTA results across all datasets, with the exception of the Num. metric on VQAv2 and the DA. metric on A-OKVQA.
Notably, when LLaVa-1.5-7B is used as the PVLM foundation for the \texttt{S-RR} module, {\scshape ViKSeR} significantly outperforms LLaVa-1.5-7B across all metrics.
In particular, on the MC. metric of the A-OKVQA dataset, {\scshape ViKSeR} achieves an almost 30\% improvement over LLaVa-1.5-7B. 
Meanwhile, {\scshape ViKSeR}, employing a gradient-free reasoning paradigm, continues to outperform task-specific methods pre-trained on the datasets.
These substantial advantages demonstrate the superiority and generalizability of {\scshape ViKSeR}'s reasoning paradigm.
Moreover, {\scshape ViKSeR}, with fewer parameters, outperforms baselines with more parameters, such as ChatGPT-4o and Molmo-72B, across all metrics except for Num. and DA. The advancements highlight {\scshape ViKSeR}'s potential in visual reasoning tasks.

On the other hand, we analyze the reasons for {\scshape ViKSeR}'s slight underperformance on the Num. metric of the VQAv2 dataset as follows:
(1){\scshape ViKSeR} has fewer model parameters, larger models, such as Molmo-72B, benefit from a larger parameter size; (2) {\scshape ViKSeR} prioritizes extracting information from key entities in the image, which may lead to a potential risk of overlooking a holistic analysis of the image. For these reasons, {\scshape ViKSeR} performs slightly worse on visual reasoning tasks involving numerical questions compared to more powerful LLMs, such as Molmo-72B. Nonetheless, we believe there is still room for improvement in {\scshape ViKSeR}'s numerical analysis performance.
Additionally, although {\scshape ViKSeR} achieves a marginally lower score than ChatGPT-4o and ChatGPT-5 on the DA. metric of A-OKVQA, the performance gap is negligible, with differences of only 1.26 and 1.94 points, respectively. These results indicate that {\scshape ViKSeR} attains comparable performance to ChatGPT-4o in visual reasoning tasks, demonstrating its competitiveness as an effective model.

\textbf{{\scshape ViKSeR} performs excellently in reasoning and understanding visual information.}
To further evaluate the reasoning performance of {\scshape ViKSeR}, we compare it with nine competitive baselines on the COLA, CREPE dataset, as shown in Table~\ref{tab:vke}. 
Specifically, {\scshape ViKSeR} outperforms existing methods, achieving a new SOTA result. 
Notably, both the COLA and CREPE datasets require a model to precisely match images with captions. Therefore, the success of {\scshape ViKSeR} highlights its capacity to accurately interpret image content and effectively extract information from key entities, enabling it to perform visual reasoning tasks with high precision.

\begin{table}[h]
\caption{Comparative performance of different methods on the Cola and CREPE datasets. The baseline scores are sourced from related studies~\cite{covlm2024,cola2023}.}
\label{tab:vke}
\begin{center}
\begin{small}
\begin{sc}
\begin{tabular}{l|c|ccr}
\toprule
\multirow{2}{*}{\textbf{Methods}} & Cola  & \multicolumn{2}{c}{CREPE} \\
\cmidrule(lr){2-2} \cmidrule(lr){3-4}
        & Acc.~(\%) & Acc.~(\%)  & R@1~(\%) \\
\midrule
MosaiCLIP    & - & - & \cellcolor{yellow!30}90.2 \\
CLIP $+$ MM-Pred  & 41.42 & 77.84 & - \\
CLIP $+$ Linear   & 30.47 & \cellcolor{yellow!30}87.35 & - \\
CoVLM  & \cellcolor{yellow!30}44.29 & - & - \\
{\scshape ViKSeR}~(ours)       & \cellcolor{lightgreen}\textbf{49.49} & \cellcolor{lightgreen}\textbf{96.68} & \cellcolor{lightgreen}\textbf{93.81} \\
\bottomrule
\end{tabular}
\end{sc}
\end{small}
\end{center}
\end{table}

\begin{figure}[h]
\begin{center}
\centerline{\includegraphics[width=\columnwidth]{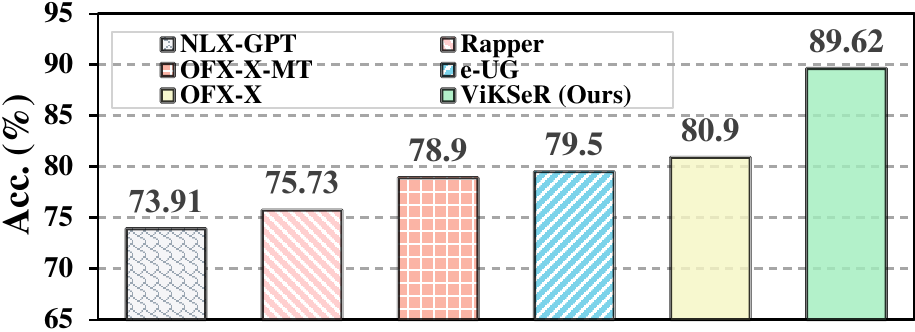}}
\caption{Comparative performance of different methods on the e-SNLI-VE dataset. The baseline scores are sourced from the official repository of e-SNLI-VE and the seminal study conducted by Chang et al.~\cite{rapper2024}.}
\label{fig:esnlive}
\end{center}
\vskip -0.2in
\end{figure}

Additionally, as demonstrated in Figure~\ref{fig:esnlive}, {\scshape ViKSeR} performs outstandingly on the e-SNLI-VE dataset.
Notably, {\scshape ViKSeR} surpasses OFX-X by 8.72\%, highlighting its effectiveness. 
The success of {\scshape ViKSeR} underscores its ability to correctly infer the visual scene present in an image based on the visual information. We attribute this advantage to the seamless collaboration between the \texttt{F-VKE} and \texttt{S-RR} modules, as well as the flexible reasoning paradigm within the \texttt{S-RR} module. As discussed in Section~\ref{subsec:SRR}, the plug-and-play nature and gradient-free design of the reasoning paradigm in the \texttt{S-RR} module endow {\scshape ViKSeR} with high generalization capabilities, enabling it to adapt to diverse visual reasoning tasks.

\textbf{{\scshape ViKSeR} demonstrates remarkable adaptability across diverse LLMs.}
To further validate the adaptability of {\scshape ViKSeR}, we conduct comparative experiments across three datasets: VQA-v2, A-OKVQA, and VizWiz. 
These experiments assess {\scshape ViKSeR} when integrated with four distinct VLM and LLM backbones: LLaVa-1.5-7B, Qwen-2.5, ChatGPT-4o, and ChatGPT-5. 
Notably, to integrate {\scshape ViKSeR} with various backbones, we utilize its original \texttt{F-VKE} module while adapting the \texttt{S-RR} module specifically for each VLM and LLM.
The results, presented in Table~\ref{tab:vikser_llm_pap}, reveal that the integration of {\scshape ViKSeR} yields significant performance improvements across all metrics and datasets for every backend. 
This outcome demonstrates that {\scshape ViKSeR} can be effectively transferred to diverse VLM and LLM backends, delivering consistent gains and showcasing its strong generalization capabilities. Notably, the performance enhancement is more substantial for LLaVa-1.5-7B than for the ChatGPT series. The particularly remarkable gain on the Ot. metric for VQA-v2 suggests that {\scshape ViKSeR}, even when paired with a comparatively weaker backbone like LLaVa-1.5-7B, can achieve results competitive with those from powerful LLM backends, thereby validating its efficacy.

\begin{table*}[h]
    \vskip -0.1in
    \centering
    \caption{Performance of {\scshape ViKSeR} with different VLM and LLM backends (LLaVa-1.5-7B, Qwen-2.5, ChatGPT-4o, and ChatGPT-5) on three datasets (VQA-v2, A-OKVQA, and VizWIz). 
    "\textbf{Base}" denotes the original performance of each model, and "\textbf{+{\scshape ViKSeR}}" denotes its performance after being enhanced by the {\scshape ViKSeR} framework.}
    \label{tab:vikser_llm_pap}
    \begin{tabular}{ll|>{\centering\arraybackslash}p{1.3cm}>{\centering\arraybackslash}p{1.3cm}>{\centering\arraybackslash}p{1.3cm}>{\centering\arraybackslash}p{1.3cm}>{\centering\arraybackslash}p{1.3cm}>{\centering\arraybackslash}p{1.3cm}>{\centering\arraybackslash}p{1.3cm}>{\centering\arraybackslash}p{1.3cm}}
        \toprule
        \multirow{2}{*}{Datasets} & \multirow{2}{*}{Metrics}  & \multicolumn{2}{c}{LLaVa-1.5-7B}  & \multicolumn{2}{c}{Qwen-2.5}  & \multicolumn{2}{c}{ChatGPT-4o} & \multicolumn{2}{c}{ ChatGPT-5}\\
        & & Base & $+${\scshape ViKSeR} & Base  & $+${\scshape ViKSeR}  & Base  & $+${\scshape ViKSeR} & Base  & $+${\scshape ViKSeR} \\
        \midrule
         
        \multirow{4}{*}{VQA-v2}    & Ov. (\%) & 74.04 & \textbf{88.74} & 64.73 & \textbf{72.45} & 86.98 & \textbf{95.52} & 88.68 & \textbf{97.13}\\
                                   & YN (\%) & 86.84 & \textbf{98.31} & 71.25 & \textbf{87.50} & 93.44 & \textbf{98.98} & 93.86 & \textbf{98.76} \\
                                   & Num. (\%) & 52.94 & \textbf{75.51} & 45.38 & \textbf{68.50} & 71.43 & \textbf{85.57} & 77.86 & \textbf{87.89} \\
                                   & Ot. (\%) & 57.89 & \textbf{86.71} & 49.78 & \textbf{79.63} & 85.94 & \textbf{93.75} & 85.33 & \textbf{96.15}\\
        \midrule
        \multirow{2}{*}{A-OKVQA}   & MC. (\%) & 62.56 & \textbf{92.51} & 56.55 & \textbf{81.35} & 90.79 & \textbf{96.83} & 91.86 & \textbf{97.26} \\
                                   & DA. (\%) & 77.38 & \textbf{88.53} & 68.93 & \textbf{79.35} & 89.79 & \textbf{95.27} & 90.47 & \textbf{95.87} \\
        \midrule
        VizWIz                 & Ov. (\%) & 57.07 & \textbf{82.52} & 55.73 & \textbf{77.86} & 79.87 & \textbf{88.68}  & 82.15 & \textbf{89.79} \\
        \bottomrule
    \end{tabular}
    \vskip -0.1in
\end{table*}

\subsection{Ablation Studies}\label{subsec:ablation studies}
We conduct various ablation experiments to evaluate the performance of each component of {\scshape ViKSeR}. Specifically, on the VQAv2 dataset, we systematically ablate the \textit{Ag-SPR} and self-reflection mechanism in the \texttt{S-RR} module to assess their impact on the performance of {\scshape ViKSeR}.
As shown in Table~\ref{tab:ablation}, on the VQAv2 dataset, ablating \textit{Ag-SPR} individually results in the least impact on overall performance.
In contrast, the simultaneous ablation of both \textit{Ag-SPR} and the self-reflection mechanism leads to the most significant performance drop.
To evaluate the performance of the \texttt{F-VKE} module, we conduct comparative experiments by ablating it and using LLaVA-1.5 and ChatGPT-4o as the LLM backbones for the \texttt{S-RR} module of {\scshape ViKSeR}.
As shown in the lower section of Table~\ref{tab:ablation}, both {\scshape ViKSeR} configurations exhibit significant performance drops without the \texttt{F-VKE} module. Notably, the LLaVA-1.5-based variant resulted in an approximately 15\% performance decline.
These results demonstrate that the fine-grained visual knowledge provided by the \texttt{F-VKE} module critically improves the model’s visual reasoning comprehension and processing. 

\begin{table}[h]
\caption{Ablation results of experiments conducted on the VQAv2 and Cola datasets. 
\begin{sc}w/o Paraphraser\end{sc} denotes the ablation of \textit{Ag-SPR}, and \begin{sc}w/o Reflection\end{sc} denotes the ablation of the self-reflection mechanism.}
\vskip 0.05in
\label{tab:ablation}
\begin{center}
\begin{small}
\begin{sc}
\begin{tabular}{p{5.2cm}|>{\centering\arraybackslash}p{1.5cm}}
\toprule
\multirow{2}{*}{\textbf{Methods}} & VQAv2  \\
\cmidrule(lr){2-2}
& Overall~(\%)\\
\midrule
{\scshape ViKSeR}~(ours)   & \cellcolor{lightgreen}\textbf{88.74} \\
w/o Paraphraser    &  81.68 \\
w/o Reflection   &  77.86 \\
w/o Paraphraser and Reflection  & 74.04\\
\bottomrule
\end{tabular}
\end{sc}
\end{small}
\end{center}
\end{table}

\begin{figure}[h]
\begin{center}
\centerline{\includegraphics[width=1\linewidth]{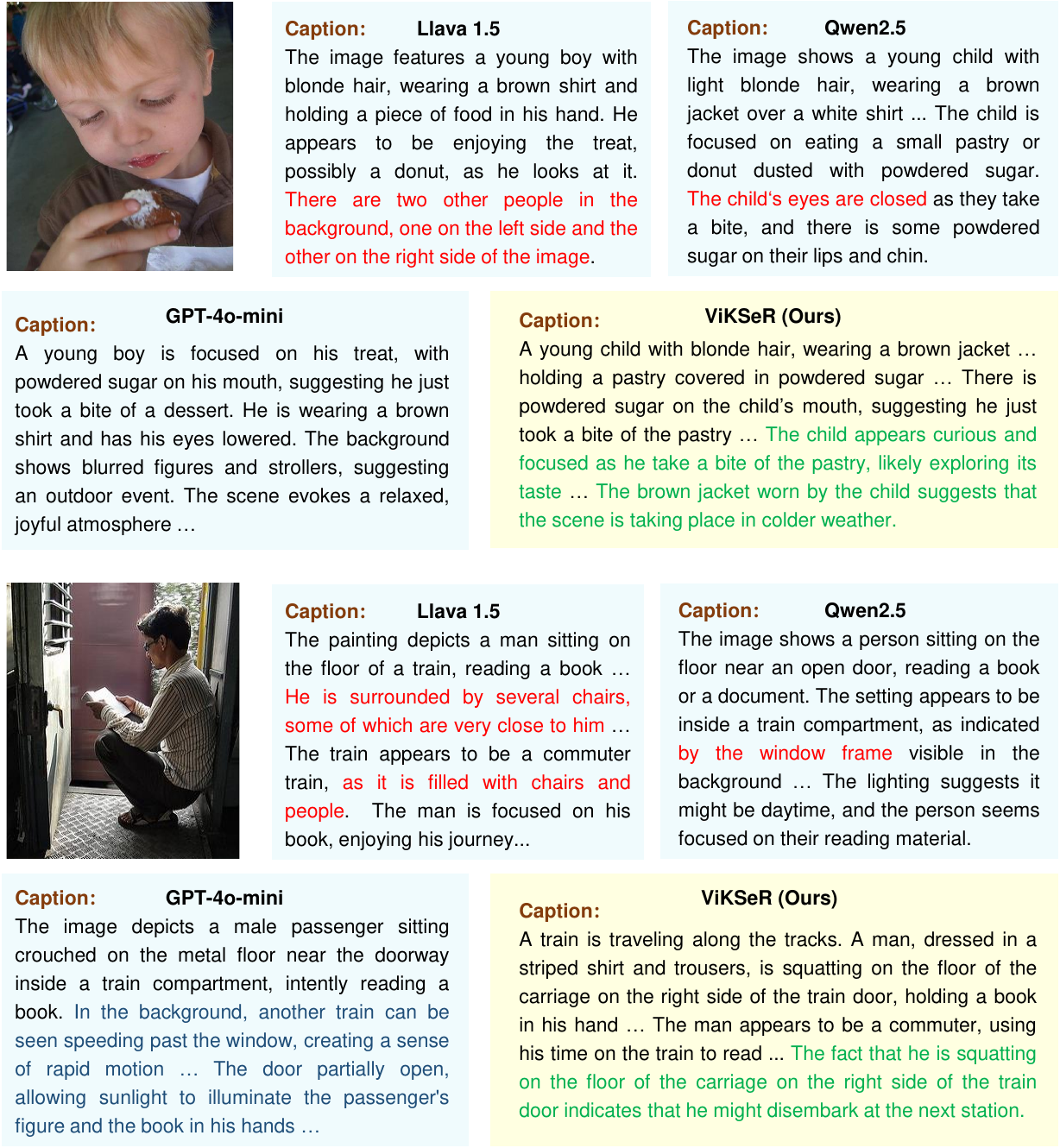}}
\caption{Performance of {\scshape ViKSeR}’s \texttt{F-VKE} module and baselines in generating captions for two images.}
\label{fig:apdcapcase1}
\end{center}
\vskip -0.2in
\end{figure}

\begin{figure}[h]
\vskip -0.1in
\begin{center}
\centerline{\includegraphics[width=1\linewidth]{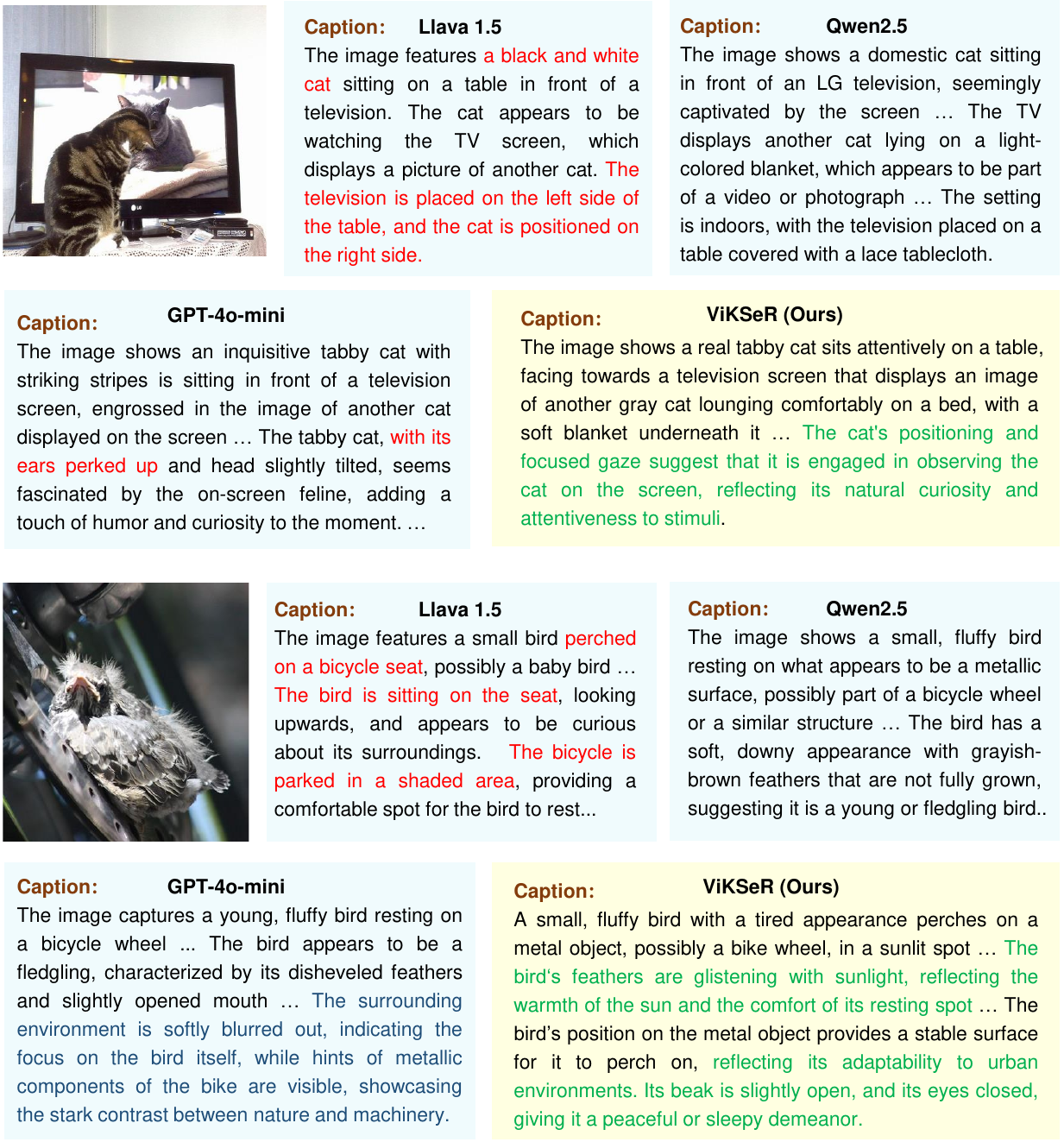}}
\caption{Performance of {\scshape ViKSeR}’s \texttt{F-VKE} module and baselines in generating captions for two images.}
\label{fig:apdcapcase2}
\end{center}
\vskip -0.3in
\end{figure}

Furthermore, to validate the efficacy of the \texttt{F-VKE} module, we conduct qualitative comparisons between {\scshape ViKSeR}'s \texttt{F-VKE} module and three baseline models (LLaVA-1.5, Qwen-2.5, and GPT-4o-mini) on detailed image caption generation across four test cases.
As shown in Figure~\ref{fig:apdcapcase1} (upper panel), both LLaVA-1.5 and Qwen-2.5 generate hallucinated content in the first case, while GPT-4o-mini produces a more accurate scene description. Crucially, {\scshape ViKSeR}'s caption demonstrates superior visual grounding through the \texttt{F-VKE} module, which not only correctly identifies image contents but also infers causal relationships embedded within the fine-grained details (e.g., "he just took a bite of the pastry" and "the scene occurs in cold weather conditions").
This pattern persists in the second case (Figure~\ref{fig:apdcapcase1}, lower panel), where LLaVA-1.5 and Qwen-2.5 again exhibit hallucination. Although GPT-4o-mini improves upon the baselines, it overemphasizes peripheral background elements. 
Fortunately, {\scshape ViKSeR}'s \texttt{F-VKE} module consistently maintains its performance advantage, delivering precise descriptions while extracting implicit causal relationships (e.g., "the subject is likely preparing to disembark at the subsequent station").
Meanwhile, in two other cases illustrated in Figure~\ref{fig:apdcapcase2}, LLaVA-1.5 continues to exhibit severe hallucinations in its generated captions. While Qwen-2.5 and GPT-4o-mini provide relatively accurate scene descriptions, the latter disproportionately emphasizes irrelevant background details.
Notably, the image captions generated by {\scshape ViKSeR}'s \texttt{F-VKE} module, which incorporates fine-grained visual knowledge, not only accurately describe the scene but also reveal causal relationships. These findings suggest that the \texttt{F-VKE} module effectively learns causal relationship inference capabilities from LLMs through knowledge distillation, enabling it to consistently produce captions enriched with fine-grained visual knowledge.

\begin{table*}[t]
\caption{Comparative performance between {\scshape ViKSeR} with CoE prompting and {\scshape ViKSeR} with CoT prompting on the VQAv2 and A-OKVQA datasets.}
\label{tab:apd coe}
\begin{center}
\begin{small}
\begin{tabular}{p{3.1cm}|>{\centering\arraybackslash}p{1.5cm}>{\centering\arraybackslash}p{1.5cm}>{\centering\arraybackslash}p{1.5cm}>{\centering\arraybackslash}p{1.5cm}|>{\centering\arraybackslash}p{1.5cm}>{\centering\arraybackslash}p{1.5cm}}
\toprule
\multirow{2}{*}{\textbf{Methods}} & \multicolumn{4}{c|}{\textbf{VQAv2}} & \multicolumn{2}{c}{\textbf{A-OKVQA}} \\
\cmidrule(lr){2-5} \cmidrule(lr){6-7}
& Ov.~(\%) & Y/N~(\%) & Num.~(\%) & Ot.~(\%) & MC.~(\%) & DA.~(\%) \\
\midrule
{\scshape ViKSeR} w/ CoE & \textbf{88.74} & \textbf{98.31} & \textbf{75.51} & \textbf{86.71} & \textbf{92.51} & \textbf{88.53} \\
{\scshape ViKSeR} w/ CoT & 71.42  & 76.92  & 61.54  & 69.23  & 79.81  & 75.76  \\
\bottomrule
\end{tabular}
\end{small}
\end{center}
\vskip -0.1in
\end{table*}

On the other hand, to systematically validate the effectiveness of the CoE prompting method, we conduct quantitative comparative experiments. 
Specifically, within the \texttt{S-RR} module of {\scshape ViKSeR}, we integrate both CoE and CoT prompting strategies and evaluate their performance on the VQAv2 and A-OKVQA datasets. 
As shown in Table~\ref{tab:apd coe}, {\scshape ViKSeR} with CoE prompting consistently outperforms the CoT-based variant across all evaluation metrics.
These results demonstrate that incorporating fine-grained visual knowledge as factual evidence enables CoE prompting to establish more reliable reasoning, leading to superior visual reasoning performance.

\begin{figure}[h]
\begin{center}
\centerline{\includegraphics[width=1\linewidth]{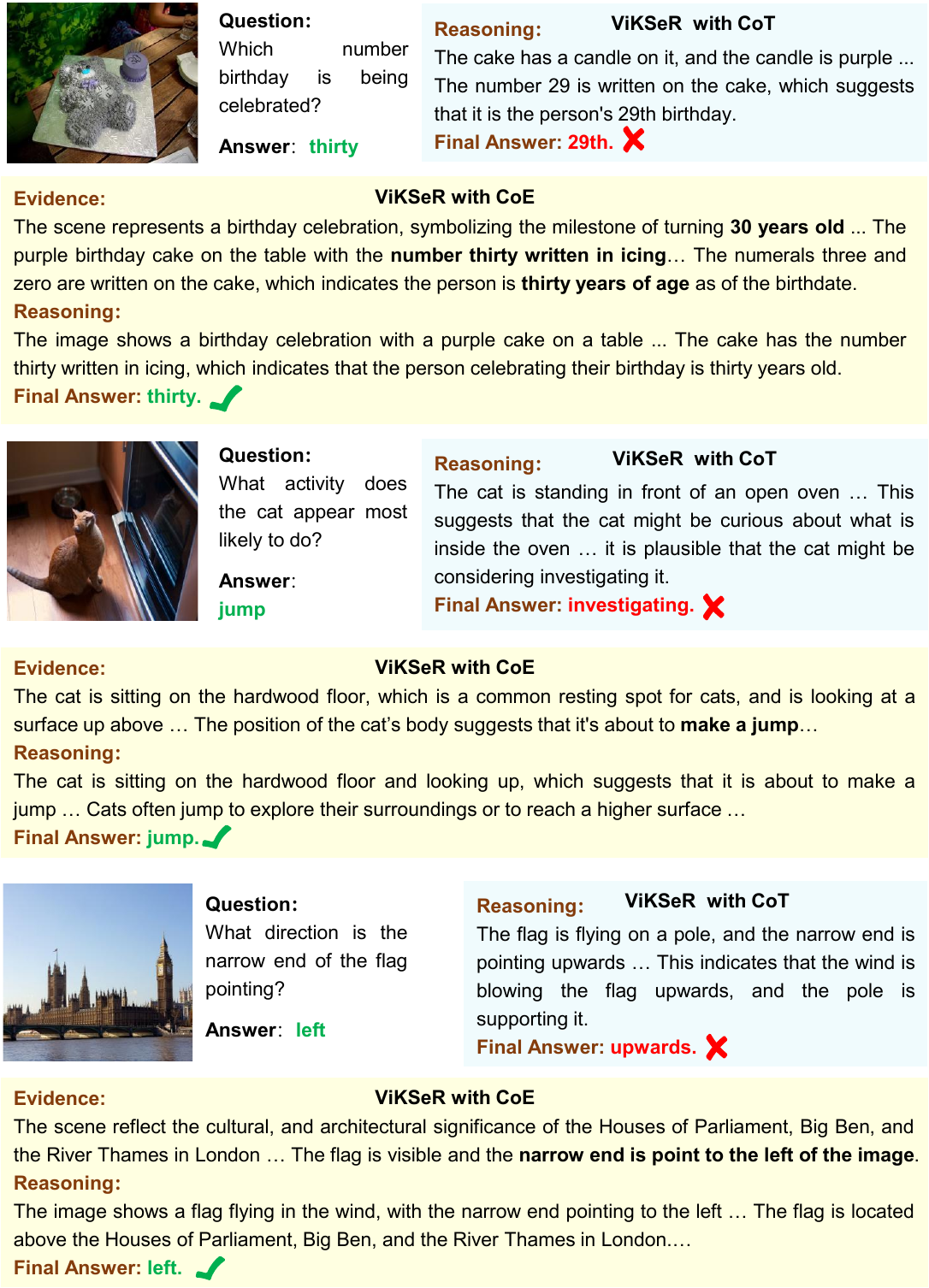}}
\caption{Comparative performance between {\scshape ViKSeR} with CoE prompting and {\scshape ViKSeR} with CoT prompting on three representative visual reasoning tasks.}
\label{fig:apdcoecases}
\end{center}
\vskip -0.2in
\end{figure}

Additionally, to further demonstrate the effectiveness of the CoE prompting method, we conduct qualitative comparative experiments. 
Specifically, we compare the performance of {\scshape ViKSeR} when integrated with CoE prompting versus CoT prompting on three representative visual reasoning tasks.
As illustrated in Figure~\ref{fig:apdcoecases}, in the first case, the CoT prompting method exhibits severe hallucination, erroneously interpreting ``30'' as ``29'', which leads to an incorrect answer.
In contrast, the CoE prompting method successfully derives the correct response by leveraging the precise \textit{Evidence} ``thirty years of age'' provided in the Evidence section, demonstrating reliable reasoning.
Similarly, in the second and third cases, the CoT prompting method fails to accurately interpret visual information and again suffers from hallucination, producing incorrect answers in both instances.
Notably, the CoE prompting method consistently performs well, relying on accurate \textit{Evidence} (``make a jump'' and ``point to the left'') and robust reasoning to correctly solve these visual reasoning tasks.
These cases demonstrate that the CoE prompting method effectively mitigates hallucination in {\scshape ViKSeR}’s reasoning while enhancing its reliability and interpretability by incorporating fine-grained visual knowledge from the \texttt{F-VKE} module as factual evidence (as discussed in Section~\ref{subsec:SRR}).

\subsection{Case Studies}\label{subsec:case study}
\begin{figure}[h]
\vskip -0.05in
\begin{center}
\centerline{\includegraphics[width=\columnwidth]{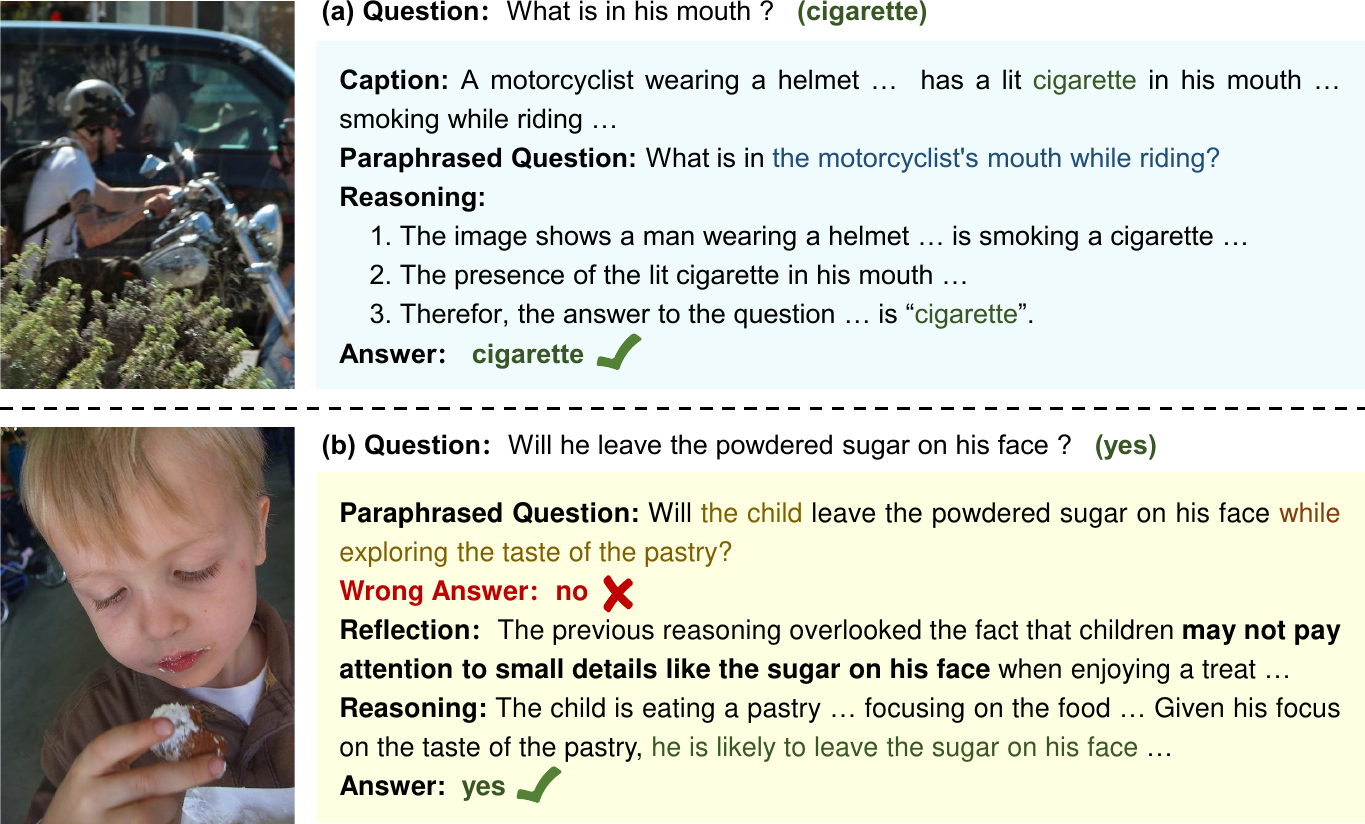}}
\vskip -0.05in
\caption{The performance of {\scshape ViKSeR} on two practical visual reasoning cases.}
\label{fig:casestudy}
\end{center}
\vskip -0.2in
\end{figure}

To further validate the effectiveness of {\scshape ViKSeR}, we conduct a qualitative analysis of its performance on two practical visual reasoning cases. 
As shown in Figure~\ref{fig:casestudy}, in case (a), {\scshape ViKSeR} identifies that the motorcyclist has a lit cigarette in his mouth while riding, by generating an image caption that contains fine-grained visual knowledge. This discovery serves as the foundation for solving the case (a).
Subsequently, {\scshape ViKSeR} utilizes the extracted visual knowledge to refine the ambiguous descriptions of the subject in the original question, resulting in a paraphrased question.
Finally, through step-by-step reasoning with evidence, {\scshape ViKSeR} infers the correct answer, ``cigarette''.
This practical visual reasoning case demonstrates {\scshape ViKSeR}'s ability to extract fine-grained visual knowledge and its proficiency in utilizing visual facts as evidence for interpretable reasoning.
Additionally, case (b) further demonstrates {\scshape ViKSeR}'s self-reflection capability. Specifically, due to inevitable reasoning errors, {\scshape ViKSeR} initially inferred an incorrect answer, ``no'', in a previous reasoning process. Fortunately, with the assistance of the self-reflection mechanism, {\scshape ViKSeR} generates a reflective message regarding its last failure, acknowledging that it had overlooked the fact that children may not pay attention to small details such as sugar on his faces while enjoying a treat. Subsequently, utilizing the reflective message, {\scshape ViKSeR} accurately guides a new round of evidence-based reasoning and ultimately infers the correct answer, ``yes''.
This practical case demonstrates {\scshape ViKSeR}'s ability to address reasoning errors through self-reflection and generate more valuable responses.

\begin{figure}[h]
\begin{center}
\centerline{\includegraphics[width=1\linewidth]{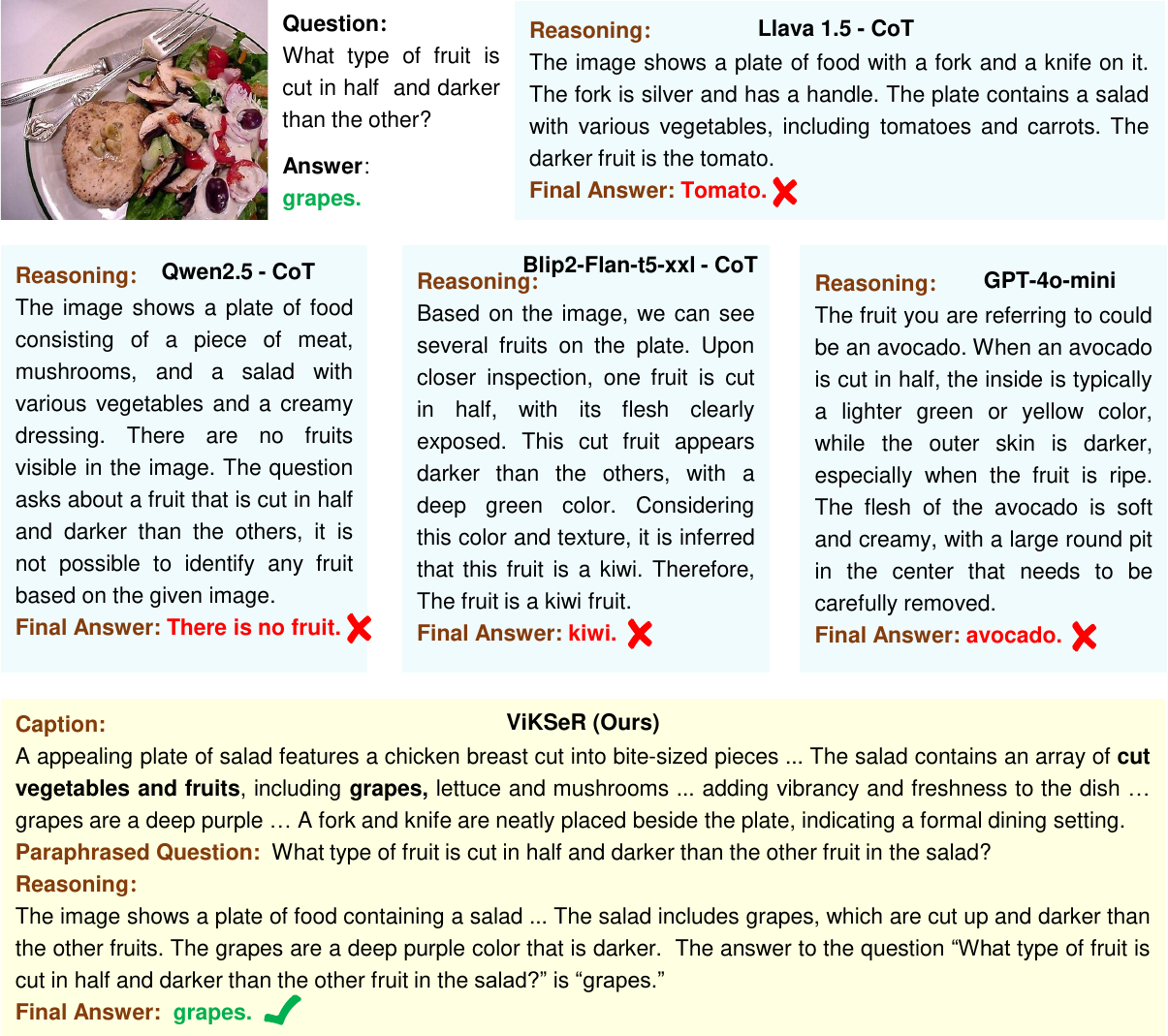}}
\caption{Performance of {\scshape ViKSeR} and baselines on a visual reasoning task involving short-answer questions.}
\label{fig:apdcase1}
\end{center}
\vskip -0.2in
\end{figure}

\begin{figure}[h]
\begin{center}
\centerline{\includegraphics[width=1\linewidth]{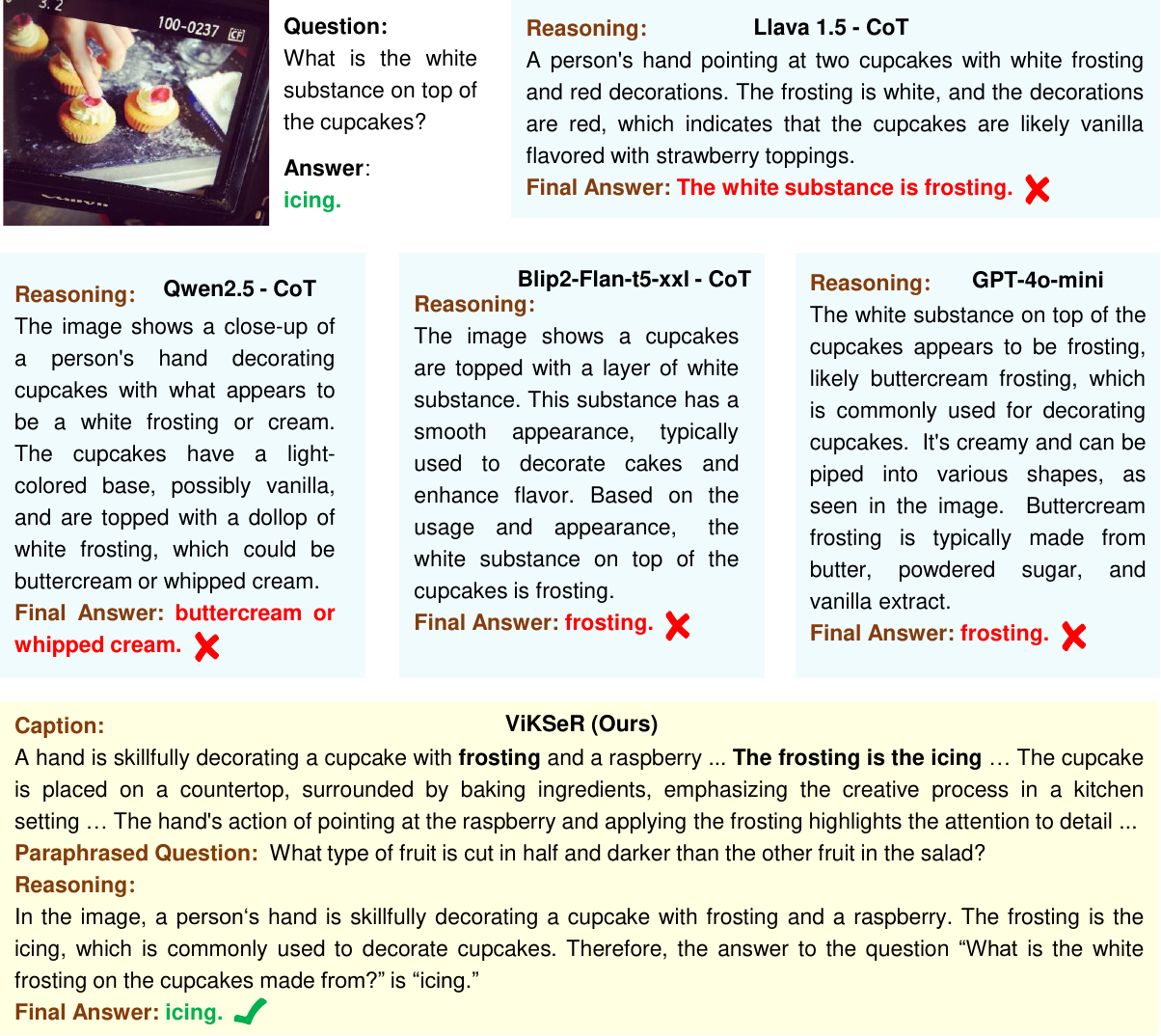}}
\caption{Performance of {\scshape ViKSeR} and baselines on a visual reasoning task involving short-answer questions.}
\label{fig:apdcase2}
\end{center}
\vskip -0.2in
\end{figure}

\begin{figure}[h]
\begin{center}
\centerline{\includegraphics[width=1\linewidth]{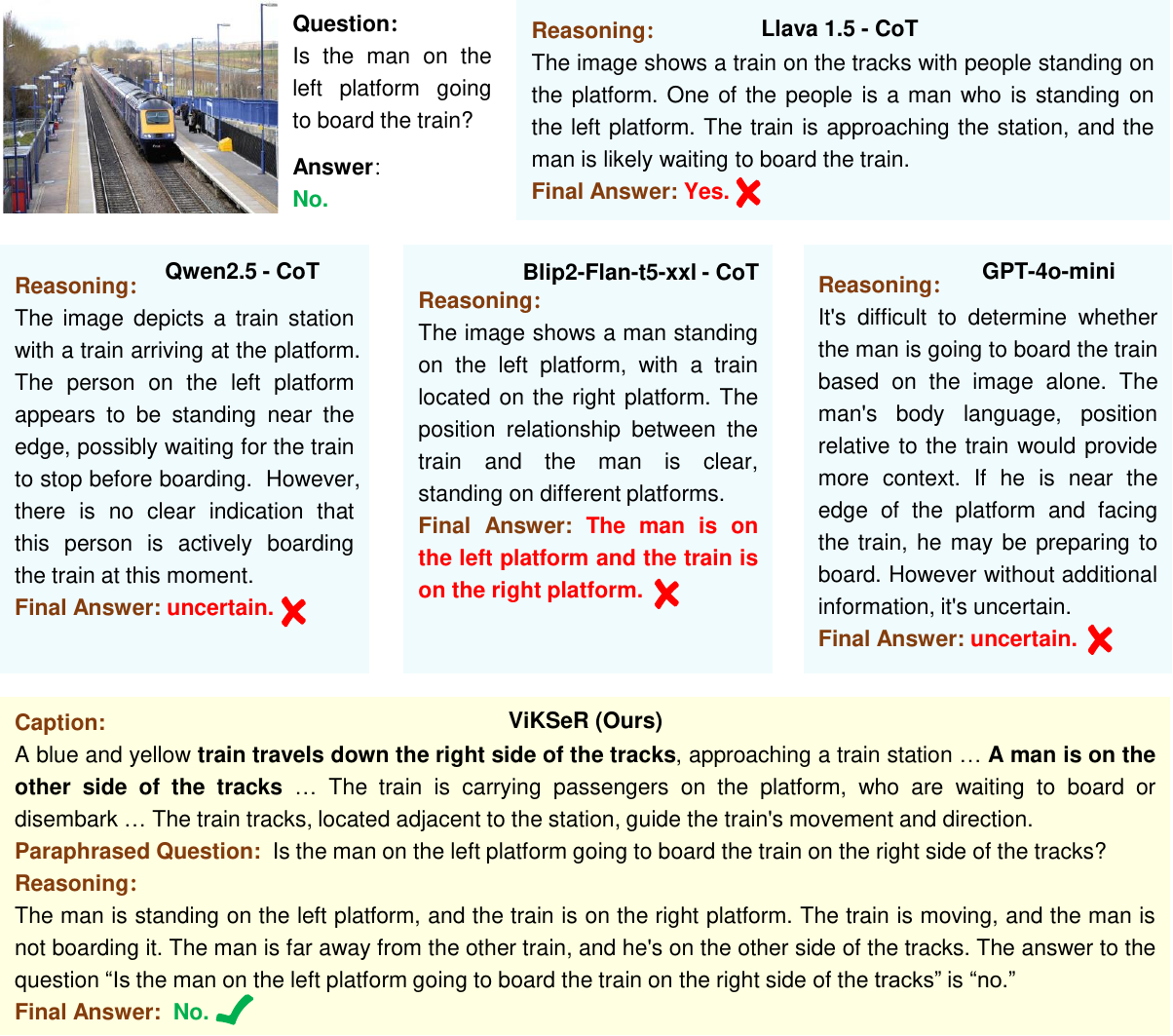}}
\caption{Performance of {\scshape ViKSeR} and baselines on a visual reasoning task involving a true/false question.}
\label{fig:apdcase3}
\end{center}
\vskip -0.2in
\end{figure}

\begin{figure}[h]
\begin{center}
\centerline{\includegraphics[width=1\linewidth]{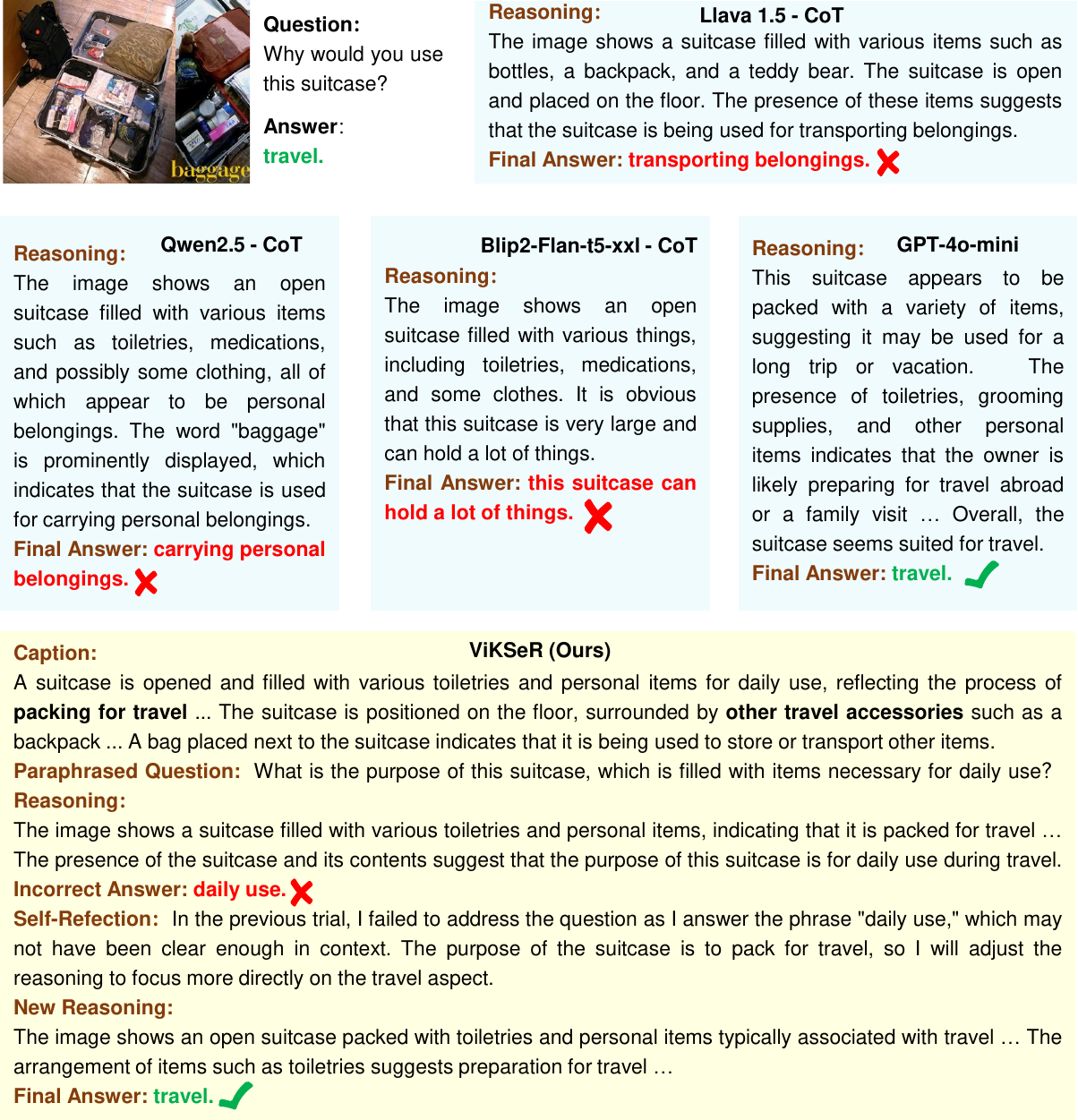}}
\caption{Performance of {\scshape ViKSeR} and baselines on a visual reasoning task involving short-answer questions.}
\label{fig:apdcase4}
\end{center}
\vskip -0.2in
\end{figure}

Moreover, we demonstrate the capabilities of {\scshape ViKSeR} on more typical visual reasoning cases. 
The baseline models selected for comparison include LLaVa-1.5, Qwen-2.5, Blip2-Flan-T5-XXL~\cite{BLIP22023}, and GPT-4o-mini.
It is worth noting that, due to the comparatively smaller parameter scales of LLaVa-1.5, Qwen-2.5, and Blip2-Flan-T5-XXL, we implement CoT prompting to augment their reasoning abilities. In contrast, for GPT-4o-mini, which possesses a substantially larger parameter size, we directly employ it for reasoning tasks without additional prompting strategies.
Detailed results are shown in Figure~\ref{fig:apdcase1}, \ref{fig:apdcase2}, \ref{fig:apdcase3} and~\ref{fig:apdcase4}.
Specifically, Figure~\ref{fig:apdcase1} and~\ref{fig:apdcase2} compare the performance of {\scshape ViKSeR} and the baselines on two visual reasoning tasks involving short-answer questions.
While all baselines fail to solve both tasks, {\scshape ViKSeR} excels by analyzing the fine-grained visual knowledge it extracts and applying correct reasoning to answer the questions.
Additionally, Figure~\ref{fig:apdcase3} illustrates the performance of {\scshape ViKSeR} and the baseline models in addressing a visual reasoning task involving true/false questions.
All baseline models fail to achieve satisfactory results in this task, especially Blip2-Flan-t5-xxl using CoT prompting, which fails to understand the question requirements and merely describes the image information in the answer.
In contrast, {\scshape ViKSeR} demonstrates exceptional performance, successfully solving the visual reasoning task with remarkable accuracy and proficiency.
Finally, Figure~\ref{fig:apdcase4} presents the performance of {\scshape ViKSeR} and the baseline models in addressing a visual reasoning task involving short-answer questions.
With the exception of GPT-4o-mini, all baseline models fail to achieve satisfactory results in this task. Notably, {\scshape ViKSeR} initially produces an incorrect result during the first round of reasoning. Fortunately, leveraging the self-reflection mechanism, {\scshape ViKSeR} identifies its mistake and re-infers the correct answer.

\section{Limitations}\label{sec:limitations}
Despite the significant improvements achieved by {\scshape ViKSeR} in visual reasoning tasks compared to existing methods, several limitations remain:
1) In \textit{Ag-VRD}, we directly employ the CoVLM model to extract visual relationships between key entities in images. However, its simplified architecture may incompletely capture complex relational patterns, suggesting that advanced VRD methods could improve performance.
2) During the knowledge distillation process for training the causal relationship analyzer $G_a$ and image caption generator $G_c$, we use ChatGPT-4o as the sole teacher model without conducting comparative experiments with other LLMs. Exploring more powerful LLMs as alternative teacher models could potentially improve performance.
3) In \textit{Ag-SR}, the self-reflection mechanism may fail to identify critical insights from prior reasoning failures during highly complex visual reasoning tasks, limiting subsequent response refinement. Mechanism optimization remains an open research direction.

\section{Conclusion}\label{sec:conclustion}
In this paper, we introduce {\scshape ViKSeR}, a new framework for visual reasoning tasks, which integrates the extraction of fine-grained visual knowledge with high interpretability and self-reinforcing reasoning capabilities.
{\scshape ViKSeR} comprises \texttt{F-VKE} and \texttt{S-RR} modules, as well as integrates advanced mechanisms to achieve superior performance.
We conduct extensive experiments on diverse public datasets.
The results demonstrate that {\scshape ViKSeR} outperforms the latest research across public datasets, achieving new SOTA results.
Our future work will focus on: 1) Exploring methods that extract finer-grained visual knowledge while performing more comprehensive visual reasoning based on overall image features. 2) Further integrating the capabilities of world models into {\scshape ViKSeR}.

\section*{Acknowledgments}
This work was supported by the National Key Research and Development Program of China (Grant No. 2022ZD0160603), and the Natural Science Foundation of Shanghai (No. 23ZR1422800).

{\appendix
\subsection{Training the  Image Caption Generator}\label{apd:trainGc}
After obtaining the analysis reports $ KA'_p $, the image caption generator $ G_c $ enriches the preliminary image description $ D $ by integrating the visual knowledge from the input image $ I $ and $ KA'_p $ to generate a detailed image caption $ C $ for $ I $.
As discussed in Section~\ref{subsubsec:VKR}, we utilize profound-ground-truth data generated by the LLM to train the $ G_c $.
Specifically, we extract pseudo-ground-truth  image caption $ C_p $ from LLM with a task-specific set of few-shot demonstrations as follows:
\begin{equation}
    KC_p = \left \{ C_p \ \middle| \ C_p \sim P_{LLM}(I, D, A'_p) \right \},
    \label{eq:KDKCp}
\end{equation}
where $ I $ denotes the input image, $ D $ represents the ground-truth preliminary image description of $ I $, and $ A’_p \in KA’_p  $ represents the pseudo-ground-truth analysis report. $ P_{LLM} $ denotes the LLM operating autoregressively. $ C_p $ denotes the the pseudo-ground-truth image caption sampled from $ P_{LLM} $, and $ KC_p $ represents the set of all $ C_p $.

Unfortunately, $ KC_p $ may contain noise and errors, adversely affecting the training of $ G_c $.
To address this, we similarly apply a post-processing mechanism to filter $ KA_p $ into $ KA'_p $.
Specifically, for each $ C_p $ in $ KC_p $, we use $F$ (the pre-trained MLLM) to assess its validity score $ S_{C_p} $ based on whether $ C_p $ correctly introduces the image. If $ S_{C_p} $ exceeds the predetermined threshold $ \tau $, the corresponding $ C_p $ is retained. The process of collecting $ KC’_p $  is formalized as follows:
\begin{equation}
    KC'_p = \left\{ C_p \ \middle| \ S_{C_p} > \tau \right\}, S_{C_p} = F\left(C_p, (I, D, A'_p)\right),
    \label{eq:KCprefine}
\end{equation}
where $ C_p $ denotes the pseudo-ground-truth image caption. $ I $ denotes the input image, $ D $ represents the ground-truth preliminary image description. $ A’_p $ represents the pseudo-ground-truth analysis report. $ \tau $ denotes the predetermined threshold. $ F $ denotes the pre-trained MLLM.
With $ KC'_p $ serving as pseudo-ground-truth data, we are able to train $ G_c $ with the distillation loss $ L_{G_c} $. The training process is formalized below:
\begin{equation}
    L_{G_c} = - \sum_{t=1}^{T} \log \left( P_{G_c} \left( C_{p,t}' \ \middle| \ C_{p,t-1}', (I, D, A'_p) \right) \right),
    \label{eq:GcKD}
\end{equation}
where $ C’_p \in KC’_p $, $ A’_p \in KA’_p $, and $ T = |C’_p| $. Finally, the trained $ G_c $ and $ G_a $ collaborate to generate a detailed image caption $ C $ for the input image $ I $.

\subsection{Implementation Details}\label{apd:Implementation Details}
\paragraph{Self Reflection}\label{apd:selfref}
As detailed in Section~\ref{subsec:SRR}, upon encountering an unsatisfactory result, Ag-SR engages in self-reflection through analyzing the paraphrased question, image caption, and reasoning trajectory to generate structured verbal reflection $ V_ref $. Specifically, $ V_ref $ pinpoints erroneous reasoning steps in the previous trajectory and suggests corrective measures, enabling more accurate subsequent inferences.  
This iterative process continues until either (a) a satisfactory result is obtained or (b) reaching the maximum iterations. It is noteworthy that we set the maximum iteration count to 3 across all experiments.

\paragraph{Knowledge Distillation from LLM}\label{apd:KDfromLLM}
As discussed in Section~\ref{subsubsec:VKR}, we utilize the knowledge reservoir embedded in LLM(e.g., ChatGPT-4o) to generate causal relationship analysis reports $ A’_p $ and detailed image captions $ C'_p $ as pseudo-ground-truth data to train causal relationship analyzer $ G_a $ and image caption generator $ G_c $.
This section details the implementation of our knowledge distillation framework.
As shown in Equation~\ref{eq:KDKAp}, given an input image $ I $  and its preliminary description $ D $, we prompt ChatGPT-4 to analyze causal relationships between key entities' behaviors in $ D $ and their outcomes inferred from $ I $, producing the raw output $ A_p $.
A post-processing mechanism then processes $ A_p $. to obtain the refined analysis reports $ A'_p $ (Equation~\ref{eq:KAprefine}).  
Subsequently, as indicated in Equation~\ref{eq:KDKCp}, we input $ I $, $ D $, and $ A'_p $ to ChatGPT-4, directing it to generate detailed image captions $ C_p $ that describes key entities, their behaviors, and causal inferences in natural language. 
Similarly, $ C_p $ undergoes the same post-processing mechanism to yield the refined image caption $ C'_p $ (Equation~\ref{eq:KCprefine}).  

We distill 500 annotated pairs of $ A'_p $ and $ C'_p $ from ChatGPT-4 to serve as pseudo-ground-truth data for training.  
As specified in Equations~\ref{eq:GaKD} and \ref{eq:GcKD}, the implementation fine-tunes LLaVA-1.5-7B using distillation losses $ \mathcal{L}_{G_a} $ and $ \mathcal{L}_{G_c} $ over 10 epochs, yielding the $ G_a $ and $ G_c $. 
The model input combines an image with a text template formed by concatenating the generation prompt with corresponding pseudo-ground-truth data.
Figures~\ref{fig:template_Ap} and \ref{fig:template_Cp} illustrate two examples of the text templates used for fine-tuning $ G_a $ and $ G_c $, respectively.  
The complete fine-tuning procedure executes on an NVIDIA V100 server with 32GB VRAM under identical initialization conditions.

\begin{figure}[h]\color{blue}
\begin{center}
\includegraphics[width=\linewidth]{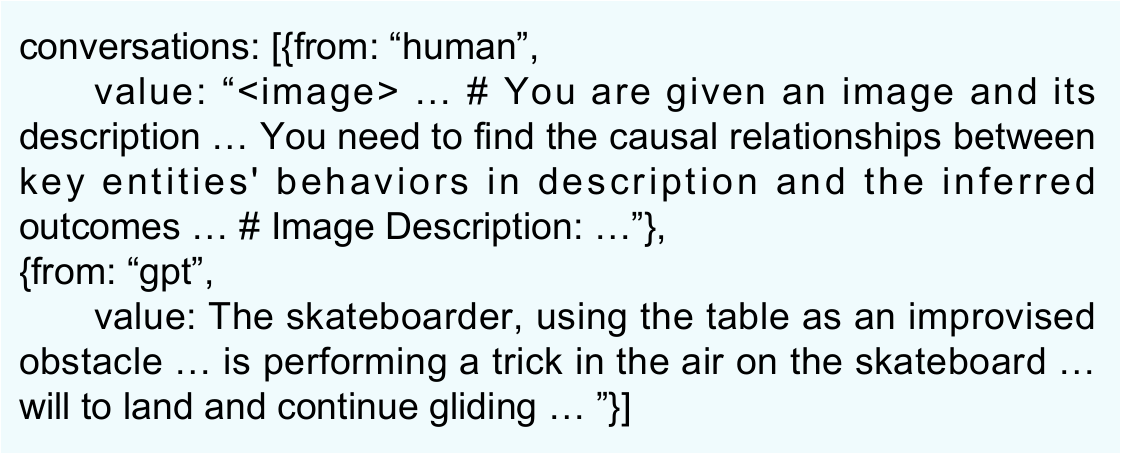}
\caption{The example of the template used for fine-turning $ G_a $.}
\label{fig:template_Ap}
\end{center}
\vskip -0.2in
\end{figure}

\begin{figure}[h]\color{blue}
\begin{center}
\includegraphics[width=\linewidth]{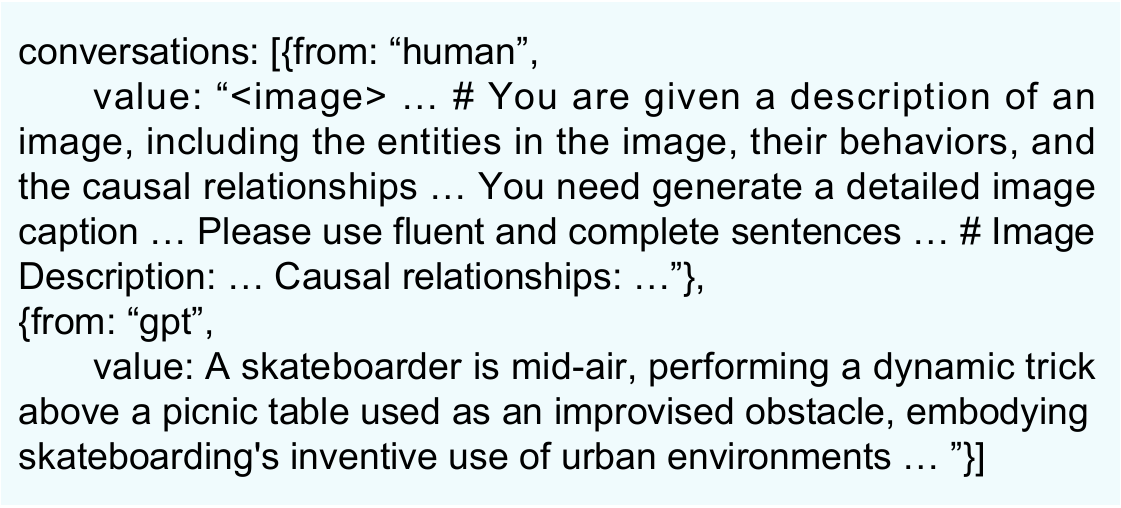}
\caption{The example of the template used for fine-turning $ C_a $.}
\label{fig:template_Cp}
\end{center}
\vskip -0.2in
\end{figure}

\paragraph{Hyperparameters}\label{apd:hyperparameters}
As introduced in Section~\ref{subsubsec:VRD}, Ag-VRD utilizes a joint entity-relation validity evaluation algorithm to compute the entity-relation joint validity scores $ S^e_r $.
Our algorithm incorporates three hyperparameters: (1) $ \gamma $  controlling the weighting of visual relationship quantity; (2) $ \alpha $ constraining the maximum number of relationships; and (3) $ \theta^e_r $ defining the validity threshold for $ S^e_r $ to identify key relationships. 
To standardize the extraction of approximately four key entities per image across all experiments, we consistently set $ \gamma = 0.1 $, $ \alpha = 4 $, and $ \theta^e_r = 0.55 $.  
Additionally, when distilling pseudo-ground-truth data from the LLM to train $ G_a $ and $ G_c $, we introduce a hyperparameter $ \tau $ in the post-processing mechanism for noise suppression. This threshold $ \tau $ acts as the validity score cutoff to filter noisy data (Equations~\ref{eq:KAprefine} and \ref{eq:KCprefine}.
Across all experiments, we set $ \tau = 0.6 $.

\subsection{Generating Reasoning from PVLMs}\label{apd:prompts}
As discussed in Section~\ref{subsec:SRR}, given the diversity of visual reasoning tasks, we propose a generalized, plug-and-play reasoning prompt paradigm to prompt PVLMs to activate the capacities of the \texttt{S-RR} module. In this section, we provide a detailed discussion of the design methodology for each prompt.

\paragraph{Specification Paraphrase}\label{apd:promptparaphrase}
After acquiring fine-grained visual knowledge from the \texttt{F-VKE} module, the \texttt{S-RR} module prompts a PVLM to paraphrase the question text that exhibits underspecification. The prompt for specification paraphrasing is shown in Figure~\ref{fig:promptparaphrase}.

\begin{figure}[h]
\begin{center}
\includegraphics[width=\linewidth]{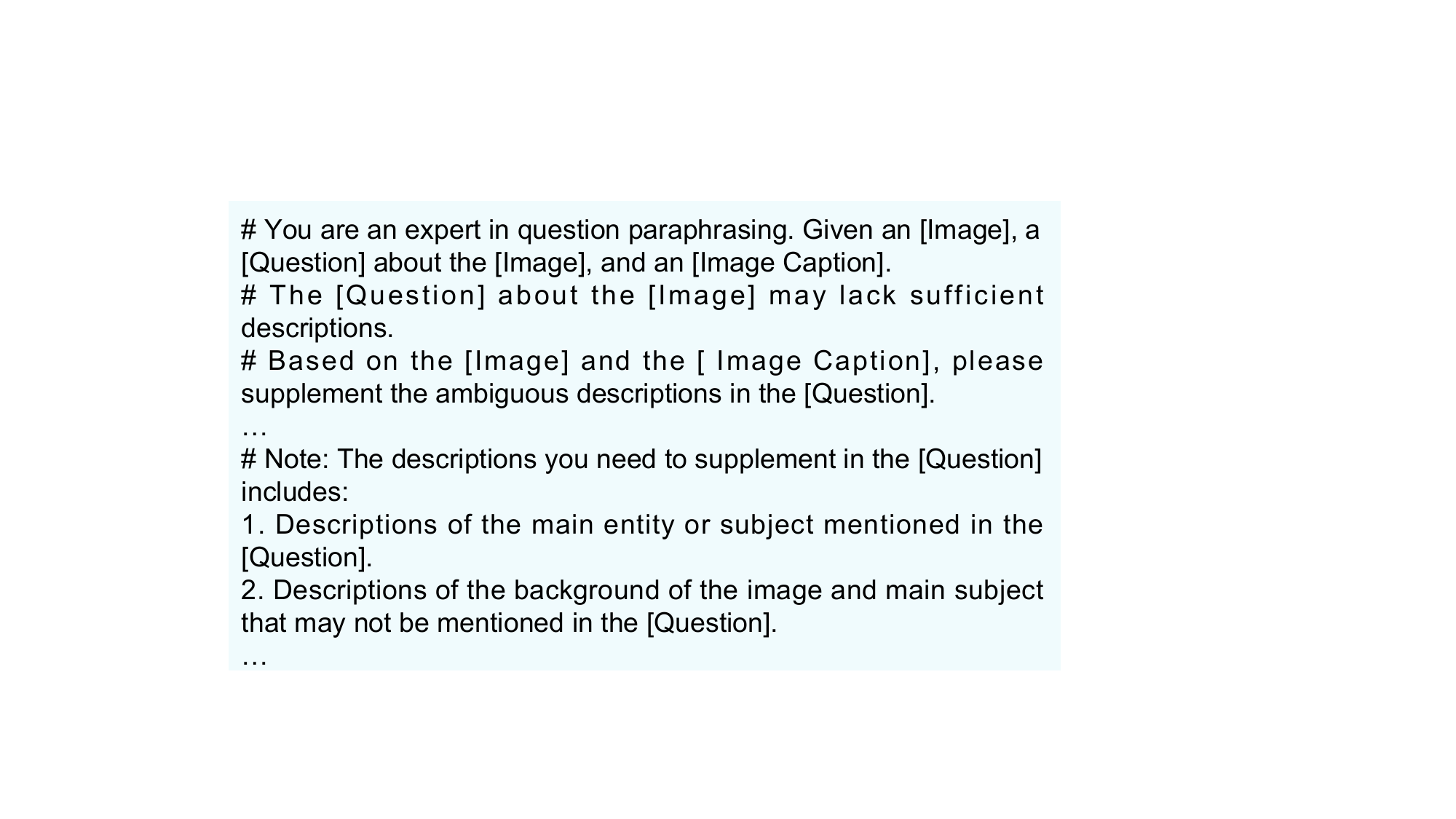}
\caption{The prompt for paraphrasing the question text with underspecification.}
\label{fig:promptparaphrase}
\end{center}
\vskip -0.2in
\end{figure}

\paragraph{CoE Prompting}\label{apd:coeprompt}
To answer the paraphrased question, the \texttt{S-RR} module employs CoE prompting to guide a PVLM in thinking step by step based on evidence and inferring a valuable response. As illustrated in Figure~\ref{fig:promptcoe}, CoE prompting first instructs the PVLM to leverage the visual information in the image and its caption to address the input question. Subsequently, the PVLM is encouraged to think step-by-step with evidence and generate both the answer and the step-by-step reasoning based on evidence.
This reasoning framework helps the PVLM focus on the relevant visual information in the image and caption, using it as evidence to support interpretable reasoning.
\begin{figure}[h]
\begin{center}
\includegraphics[width=\linewidth]{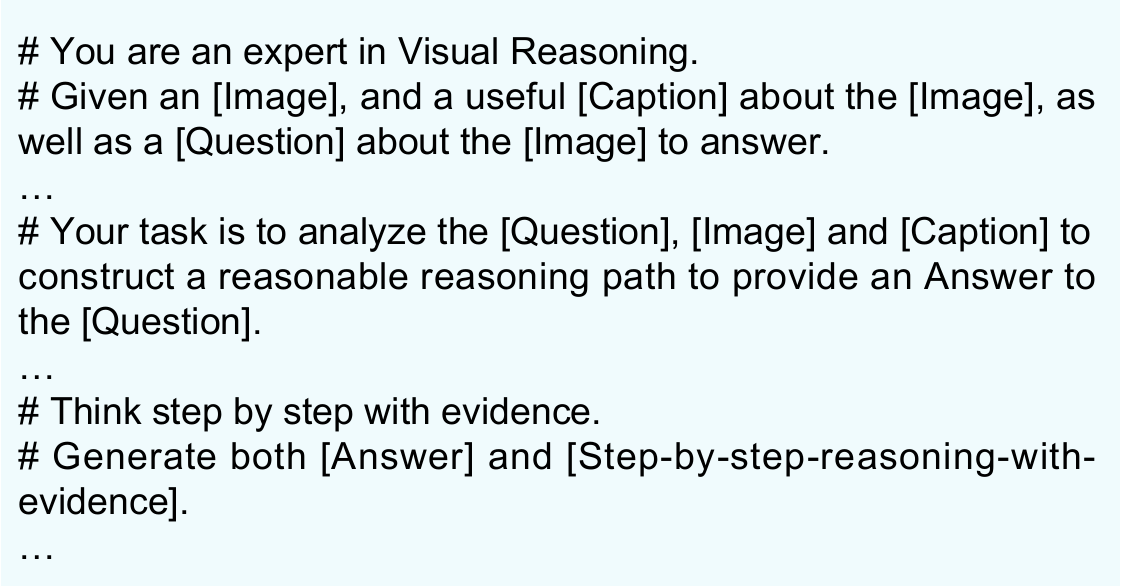}
\caption{The prompt for the CoE prompting technique.}
\label{fig:promptcoe}
\end{center}
\vskip -0.2in
\end{figure}

\paragraph{Self Reflection}\label{apd:reflection}
With the aim of addressing low-quality responses, the \texttt{S-RR} module prompts a PVLM to self-reflect on past failures and seek more valuable responses.
As illustrated in Figure~\ref{fig:promptreflection}, the PVLM is tasked with analyzing the reasoning trajectory of past failures, reflecting on their causes, and formulating a high-level plan to prevent similar failures in the future.
Finally, the PVLM will derive a more valuable answer based on the reflection information and the high-level plan.
\begin{figure}[h]
\begin{center}
\includegraphics[width=\linewidth]{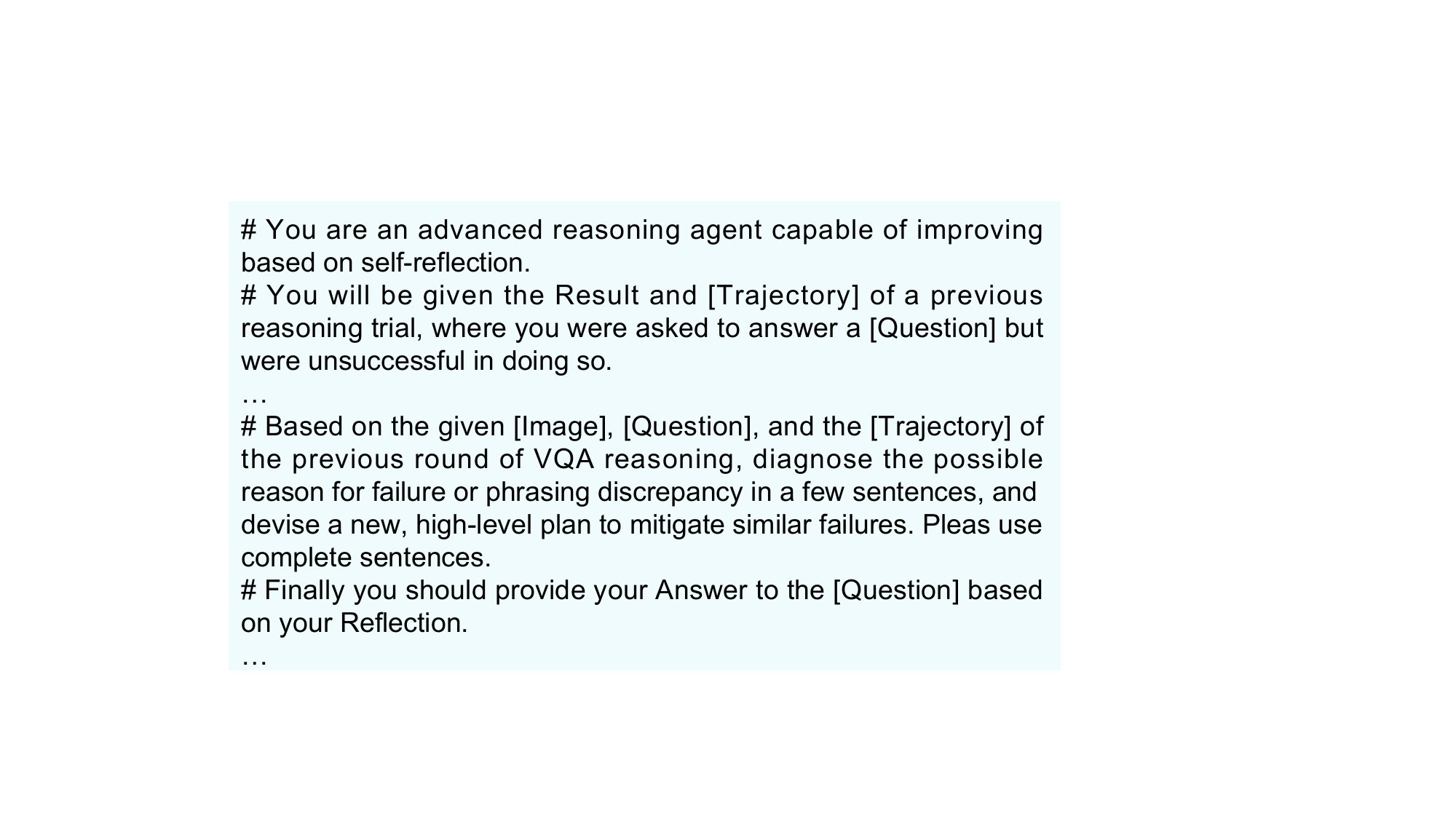}
\caption{The prompt for the self-reflection mechanism.}
\label{fig:promptreflection}
\end{center}
\vskip -0.2in
\end{figure}
}
\bibliographystyle{IEEEtran}
\bibliography{references}

\begin{thebibliography}{10}
\providecommand{\url}[1]{#1}
\csname url@samestyle\endcsname
\providecommand{\newblock}{\relax}
\providecommand{\bibinfo}[2]{#2}
\providecommand{\BIBentrySTDinterwordspacing}{\spaceskip=0pt\relax}
\providecommand{\BIBentryALTinterwordstretchfactor}{4}
\providecommand{\BIBentryALTinterwordspacing}{\spaceskip=\fontdimen2\font plus
\BIBentryALTinterwordstretchfactor\fontdimen3\font minus \fontdimen4\font\relax}
\providecommand{\BIBforeignlanguage}[2]{{%
\expandafter\ifx\csname l@#1\endcsname\relax
\typeout{** WARNING: IEEEtran.bst: No hyphenation pattern has been}%
\typeout{** loaded for the language `#1'. Using the pattern for}%
\typeout{** the default language instead.}%
\else
\language=\csname l@#1\endcsname
\fi
#2}}
\providecommand{\BIBdecl}{\relax}
\BIBdecl

\bibitem{paliGemma2024}
A.~Steiner, A.~S. Pinto, M.~Tschannen, D.~Keysers, X.~Wang, Y.~Bitton, A.~Gritsenko, M.~Minderer, A.~Sherbondy, S.~Long \emph{et~al.}, ``Paligemma 2: A family of versatile vlms for transfer,'' \emph{arXiv preprint arXiv:2412.03555}, 2024.

\bibitem{llava12023}
H.~Liu, C.~Li, Q.~Wu, and Y.~J. Lee, ``Visual instruction tuning,'' 2023.

\bibitem{VRinVLM2021}
A.~Kamath, M.~Singh, Y.~LeCun, G.~Synnaeve, I.~Misra, and N.~Carion, ``Mdetr-modulated detection for end-to-end multi-modal understanding,'' in \emph{Proceedings of the IEEE/CVF international conference on computer vision}, 2021, pp. 1780--1790.

\bibitem{smola2024}
J.~Wu, X.~Hu, Y.~Wang, B.~Pang, and R.~Soricut, ``Omni-smola: Boosting generalist multimodal models with soft mixture of low-rank experts,'' in \emph{Proceedings of the IEEE/CVF Conference on Computer Vision and Pattern Recognition}, 2024, pp. 14\,205--14\,215.

\bibitem{cola2023}
A.~Ray, F.~Radenovic, A.~Dubey, B.~Plummer, R.~Krishna, and K.~Saenko, ``Cola: A benchmark for compositional text-to-image retrieval,'' \emph{Advances in Neural Information Processing Systems}, vol.~36, 2024.

\bibitem{VPCVR2023}
T.~Gupta and A.~Kembhavi, ``Visual programming: Compositional visual reasoning without training,'' in \emph{Proceedings of the IEEE/CVF Conference on Computer Vision and Pattern Recognition (CVPR)}, June 2023, pp. 14\,953--14\,962.

\bibitem{visualreasoning2021}
F.~He, Y.~Wang, X.~Miao, and X.~Sun, ``Interpretable visual reasoning: A survey,'' \emph{Image and Vision Computing}, vol. 112, p. 104194, 2021.

\bibitem{rapper2024}
\BIBentryALTinterwordspacing
K.-P. Chang, C.-P. Huang, W.-Y. Cheng, F.-E. Yang, C.-Y. Wang, Y.-H. Lai, and Y.-C.~F. Wang, ``Rapper: Reinforced rationale-prompted paradigm for natural language explanation in visual question answering,'' in \emph{Proceedings of the 2024 International Conference on Learning Representations (ICLR)}.\hskip 1em plus 0.5em minus 0.4em\relax OpenReview, 2024. [Online]. Available: \url{https://openreview.net/forum?id=bshfchPM9H}
\BIBentrySTDinterwordspacing

\bibitem{repare2024}
\BIBentryALTinterwordspacing
A.~Prasad, E.~Stengel-Eskin, and M.~Bansal, ``Rephrase, augment, reason: Visual grounding of questions for vision-language models,'' in \emph{The Twelfth International Conference on Learning Representations}, 2024. [Online]. Available: \url{https://openreview.net/forum?id=L4nOxziGf9}
\BIBentrySTDinterwordspacing

\bibitem{VR2023}
Y.~Zhang, S.~Chen, and Q.~Zhao, ``Toward multi-granularity decision-making: Explicit visual reasoning with hierarchical knowledge,'' in \emph{Proceedings of the IEEE/CVF International Conference on Computer Vision}, 2023, pp. 2573--2583.

\bibitem{molmo72b2024}
M.~Deitke, C.~Clark, S.~Lee, R.~Tripathi, Y.~Yang, J.~S. Park, M.~Salehi, N.~Muennighoff, K.~Lo, L.~Soldaini \emph{et~al.}, ``Molmo and pixmo: Open weights and open data for state-of-the-art multimodal models,'' \emph{arXiv preprint arXiv:2409.17146}, 2024.

\bibitem{vlmvqamedical2023}
Y.~Bazi, M.~M.~A. Rahhal, L.~Bashmal, and M.~Zuair, ``Vision--language model for visual question answering in medical imagery,'' \emph{Bioengineering}, vol.~10, no.~3, p. 380, 2023.

\bibitem{lingoqa2024}
A.-M. Marcu, L.~Chen, J.~H{\"u}nermann, A.~Karnsund, B.~Hanotte, P.~Chidananda, S.~Nair, V.~Badrinarayanan, A.~Kendall, J.~Shotton \emph{et~al.}, ``Lingoqa: Visual question answering for autonomous driving,'' in \emph{European Conference on Computer Vision}.\hskip 1em plus 0.5em minus 0.4em\relax Springer, 2024, pp. 252--269.

\bibitem{VQAandVR2022}
R.~Y. Zakari, J.~W. Owusu, H.~Wang, K.~Qin, Z.~K. Lawal, and Y.~Dong, ``Vqa and visual reasoning: An overview of recent datasets, methods and challenges,'' \emph{arXiv preprint arXiv:2212.13296}, 2022.

\bibitem{specification2023}
S.~Pezzelle, ``Dealing with semantic underspecification in multimodal nlp,'' \emph{arXiv preprint arXiv:2306.05240}, 2023.

\bibitem{openvik2024}
H.~Cui, X.~Fang, Z.~Zhang, R.~Xu, X.~Kan, X.~Liu, Y.~Yu, M.~Li, Y.~Song, and C.~Yang, ``Open visual knowledge extraction via relation-oriented multimodality model prompting,'' \emph{Advances in Neural Information Processing Systems}, vol.~36, 2024.

\bibitem{ivlr2024}
C.~Yang, R.~Xu, Y.~Guo, P.~Huang, Y.~Chen, W.~Ding, Z.~Wang, and H.~Zhou, ``Improving vision-and-language reasoning via spatial relations modeling,'' in \emph{Proceedings of the IEEE/CVF Winter Conference on Applications of Computer Vision (WACV)}, January 2024, pp. 769--778.

\bibitem{PV2TEA2023}
H.~Cui, R.~Lin, N.~Zalmout, C.~Zhang, J.~Shang, C.~Yang, and X.~Li, ``Pv2tea: Patching visual modality to textual-established information extraction,'' \emph{arXiv preprint arXiv:2306.01016}, 2023.

\bibitem{DLimgcap2018}
W.~Jiang, L.~Ma, Y.-G. Jiang, W.~Liu, and T.~Zhang, ``Recurrent fusion network for image captioning,'' in \emph{Proceedings of the European Conference on Computer Vision (ECCV)}, September 2018.

\bibitem{DLimgcap2017}
T.~Yao, Y.~Pan, Y.~Li, Z.~Qiu, and T.~Mei, ``Boosting image captioning with attributes,'' in \emph{Proceedings of the IEEE International Conference on Computer Vision (ICCV)}, Oct 2017.

\bibitem{promptcap2023}
Y.~Hu, H.~Hua, Z.~Yang, W.~Shi, N.~A. Smith, and J.~Luo, ``Promptcap: Prompt-guided image captioning for vqa with gpt-3,'' in \emph{Proceedings of the IEEE/CVF International Conference on Computer Vision (ICCV)}, October 2023, pp. 2963--2975.

\bibitem{ISCimgcap2022}
Z.~Fang, J.~Wang, X.~Hu, L.~Liang, Z.~Gan, L.~Wang, Y.~Yang, and Z.~Liu, ``Injecting semantic concepts into end-to-end image captioning,'' in \emph{Proceedings of the IEEE/CVF Conference on Computer Vision and Pattern Recognition (CVPR)}, June 2022, pp. 18\,009--18\,019.

\bibitem{VCHG2023}
Z.~Li, X.~Zhu, X.~Zhang, Z.~Zhang, and Z.~Lei, ``Visual commonsense based heterogeneous graph contrastive learning,'' \emph{arXiv preprint arXiv:2311.06553}, 2023.

\bibitem{covlm2024}
\BIBentryALTinterwordspacing
J.~Li, D.~Chen, Y.~Hong, Z.~Chen, P.~Chen, Y.~Shen, and C.~Gan, ``Covlm: Composing visual entities and relationships in large language models via communicative decoding,'' in \emph{The Twelfth International Conference on Learning Representations, {ICLR} 2024, Vienna, Austria, May 7-11, 2024}.\hskip 1em plus 0.5em minus 0.4em\relax OpenReview.net, 2024. [Online]. Available: \url{https://openreview.net/forum?id=PHGxChm1l5}
\BIBentrySTDinterwordspacing

\bibitem{BKG2GSG2020}
A.~Zareian, S.~Karaman, and S.-F. Chang, ``Bridging knowledge graphs to generate scene graphs,'' in \emph{Computer Vision -- ECCV 2020}, A.~Vedaldi, H.~Bischof, T.~Brox, and J.-M. Frahm, Eds.\hskip 1em plus 0.5em minus 0.4em\relax Cham: Springer International Publishing, 2020, pp. 606--623.

\bibitem{cotrainfewshot2022}
H.~Lang, M.~N. Agrawal, Y.~Kim, and D.~Sontag, ``Co-training improves prompt-based learning for large language models,'' in \emph{International Conference on Machine Learning}.\hskip 1em plus 0.5em minus 0.4em\relax PMLR, 2022, pp. 11\,985--12\,003.

\bibitem{llm0shottrain2024}
Y.~Yu, Y.~Zhuang, J.~Zhang, Y.~Meng, A.~J. Ratner, R.~Krishna, J.~Shen, and C.~Zhang, ``Large language model as attributed training data generator: A tale of diversity and bias,'' \emph{Advances in Neural Information Processing Systems}, vol.~36, 2024.

\bibitem{llm0shot2022}
T.~Kojima, S.~S. Gu, M.~Reid, Y.~Matsuo, and Y.~Iwasawa, ``Large language models are zero-shot reasoners,'' \emph{Advances in neural information processing systems}, vol.~35, pp. 22\,199--22\,213, 2022.

\bibitem{palix2022}
X.~Chen, X.~Wang, S.~Changpinyo, A.~Piergiovanni, P.~Padlewski, D.~Salz, S.~Goodman, A.~Grycner, B.~Mustafa, L.~Beyer \emph{et~al.}, ``Pali: A jointly-scaled multilingual language-image model,'' \emph{arXiv preprint arXiv:2209.06794}, 2022.

\bibitem{esnlive2021}
M.~Kayser, O.-M. Camburu, L.~Salewski, C.~Emde, V.~Do, Z.~Akata, and T.~Lukasiewicz, ``e-vil: A dataset and benchmark for natural language explanations in vision-language tasks,'' in \emph{Proceedings of the IEEE/CVF international conference on computer vision}, 2021, pp. 1244--1254.

\bibitem{nlxgpt2022}
\BIBentryALTinterwordspacing
F.~Sammani, T.~Mukherjee, and N.~Deligiannis, ``Nlx-gpt: A model for natural language explanations in vision and vision-language tasks,'' \emph{CoRR}, vol. abs/2203.05081, 2022. [Online]. Available: \url{https://doi.org/10.48550/arXiv.2203.05081}
\BIBentrySTDinterwordspacing

\bibitem{spatialRelation2024}
C.~Yang, R.~Xu, Y.~Guo, P.~Huang, Y.~Chen, W.~Ding, Z.~Wang, and H.~Zhou, ``Improving vision-and-language reasoning via spatial relations modeling,'' in \emph{Proceedings of the IEEE/CVF Winter Conference on Applications of Computer Vision (WACV)}, January 2024, pp. 769--778.

\bibitem{gpt42023}
J.~Achiam, S.~Adler, S.~Agarwal, L.~Ahmad, I.~Akkaya, F.~L. Aleman, D.~Almeida, J.~Altenschmidt, S.~Altman, S.~Anadkat \emph{et~al.}, ``Gpt-4 technical report,'' \emph{arXiv preprint arXiv:2303.08774}, 2023.

\bibitem{COT2022}
\BIBentryALTinterwordspacing
J.~Wei, X.~Wang, D.~Schuurmans, M.~Bosma, brian ichter, F.~Xia, E.~H. Chi, Q.~V. Le, and D.~Zhou, ``Chain of thought prompting elicits reasoning in large language models,'' in \emph{Advances in Neural Information Processing Systems}, A.~H. Oh, A.~Agarwal, D.~Belgrave, and K.~Cho, Eds., 2022. [Online]. Available: \url{https://openreview.net/forum?id=_VjQlMeSB_J}
\BIBentrySTDinterwordspacing

\bibitem{reflexion2023}
N.~Shinn, F.~Cassano, A.~Gopinath, K.~Narasimhan, and S.~Yao, ``Reflexion: Language agents with verbal reinforcement learning,'' \emph{Advances in Neural Information Processing Systems}, vol.~36, 2024.

\bibitem{vqav22017}
Y.~Goyal, T.~Khot, D.~Summers-Stay, D.~Batra, and D.~Parikh, ``Making the v in vqa matter: Elevating the role of image understanding in visual question answering,'' in \emph{Proceedings of the IEEE conference on computer vision and pattern recognition}, 2017, pp. 6904--6913.

\bibitem{aokvqa2022}
D.~Schwenk, A.~Khandelwal, C.~Clark, K.~Marino, and R.~Mottaghi, ``A-okvqa: A benchmark for visual question answering using world knowledge,'' in \emph{European conference on computer vision}.\hskip 1em plus 0.5em minus 0.4em\relax Springer, 2022, pp. 146--162.

\bibitem{vizwiz2018}
D.~Gurari, Q.~Li, A.~J. Stangl, A.~Guo, C.~Lin, K.~Grauman, J.~Luo, and J.~P. Bigham, ``Vizwiz grand challenge: Answering visual questions from blind people,'' in \emph{Proceedings of the IEEE Conference on Computer Vision and Pattern Recognition (CVPR)}, June 2018.

\bibitem{crepe2023}
\BIBentryALTinterwordspacing
Z.~Ma, J.~Hong, M.~O. Gul, M.~Gandhi, I.~Gao, and R.~Krishna, ``@ crepe: Can vision-language foundation models reason compositionally?'' in \emph{Proceedings of the 2023 IEEE/CVF Conference on Computer Vision and Pattern Recognition}.\hskip 1em plus 0.5em minus 0.4em\relax IEEE, 2023, pp. 10\,910--10\,921. [Online]. Available: \url{https://doi.org/10.1109/CVPR52729.2023.01050}
\BIBentrySTDinterwordspacing

\bibitem{qwen2.5-VL}
\BIBentryALTinterwordspacing
Q.~Team, ``Qwen2.5-vl,'' January 2025. [Online]. Available: \url{https://qwenlm.github.io/blog/qwen2.5-vl/}
\BIBentrySTDinterwordspacing

\bibitem{4omini2024}
G.~OpenAI, ``4o mini: Advancing cost-efficient intelligence, 2024,'' \emph{URL: https://openai. com/index/gpt-4o-mini-advancing-cost-efficient-intelligence}, 2024.

\bibitem{gpt4o2024}
{OpenAI}, ``Hello gpt-4o,'' \url{https://openai.com/index/hello-gpt-4o/}, 2024, accessed 15 March 2025.

\bibitem{gpt52025}
------, ``Introducing gpt-5,'' \url{https://openai.com/zh-Hans-CN/index/introducing-gpt-5/}, 2025, published on 2025-08-09. Accessed: 2025-08-22.

\bibitem{LocVLM2024}
K.~Ranasinghe, S.~N. Shukla, O.~Poursaeed, M.~S. Ryoo, and T.-Y. Lin, ``Learning to localize objects improves spatial reasoning in visual-llms,'' in \emph{Proceedings of the IEEE/CVF Conference on Computer Vision and Pattern Recognition}, 2024, pp. 12\,977--12\,987.

\bibitem{HiMix2025}
X.~Zhang, D.~Li, B.~Liu, Z.~Bao, Y.~Zhou, B.~Yang, Z.~Liu, Y.~Zhong, Z.~Zhao, and T.~Yuan, ``Himix: Reducing computational complexity in large vision-language models,'' \emph{arXiv preprint arXiv:2501.10318}, 2025.

\bibitem{CIBi2024}
Y.~Liu, G.~Bai, L.~Chenji, S.~Li, Z.~Zhang, R.~Liu, and W.~Guo, ``Eliminating the language bias for visual question answering with fine-grained causal intervention,'' in \emph{2024 IEEE International Conference on Multimedia and Expo (ICME)}.\hskip 1em plus 0.5em minus 0.4em\relax IEEE, 2024, pp. 1--6.

\bibitem{MAVL2024}
Z.~Hu, P.~Yang, B.~Li, and Z.~Wang, ``Multi-agents based on large language models for knowledge-based visual question answering,'' \emph{arXiv preprint arXiv:2412.18351}, 2024.

\bibitem{DIETCOKE2024}
M.~Li, H.~Li, Z.~Du, and B.~Li, ``Diversify, rationalize, and combine: Ensembling multiple qa strategies for zero-shot knowledge-based vqa,'' \emph{arXiv preprint arXiv:2406.12746}, 2024.

\bibitem{ofx2022}
\BIBentryALTinterwordspacing
B.~Plüster, J.~Ambsdorf, L.~Braach, J.~H. Lee, and S.~Wermter, ``Harnessing the power of multi-task pretraining for ground-truth level natural language explanations,'' \emph{CoRR}, vol. abs/2212.04231, 2022. [Online]. Available: \url{https://doi.org/10.48550/arXiv.2212.04231}
\BIBentrySTDinterwordspacing

\bibitem{mosai2023}
H.~Singh, P.~Zhang, Q.~Wang, M.~Wang, W.~Xiong, J.~Du, and Y.~Chen, ``Coarse-to-fine contrastive learning in image-text-graph space for improved vision-language compositionality,'' \emph{arXiv preprint arXiv:2305.13812}, 2023.

\bibitem{BLIP22023}
J.~Li, D.~Li, S.~Savarese, and S.~Hoi, ``Blip-2: Bootstrapping language-image pre-training with frozen image encoders and large language models,'' in \emph{International conference on machine learning}.\hskip 1em plus 0.5em minus 0.4em\relax PMLR, 2023, pp. 19\,730--19\,742.

\end{thebibliography}

\vfill

\end{document}